\documentclass{llncs}
\usepackage{epsf,epsfig,makeidx}  % allows for indexgeneration
\def\xdlzo{{\cal MTALC}_{0,1}({\cal D}_{\combalg})}
\def\wrt{\mbox{w.r.t.}}
\def\gis{{\mbox{\em GIS}}}
\def\alcd{{\cal ALC}({\cal D})}
\def\inferred{\otimes}
\def\temp{{\mbox{\em Temp}}}
\def\pcsfc{{\mbox{\em PcS4c+()}}}
\def\queue{{\mbox{\em Queue}}}
\def\cdalg{{\cal CDA}}
\def\roalg{{\cal ROA}}
\def\combalg{c{\cal COA}}
\def\fctroatocdaontw{\mbox{roa-to-cda12}}
\def\fctroatocdaonth{\mbox{roa-to-cda13}}
\def\fctroatocdatwth{\mbox{roa-to-cda23}}
\def\cdatoroa{\mbox{$\cdalg$-to-$\roalg$}}
\def\roatocda{\mbox{$\roalg$-to-$\cdalg$}}
\def\roair{\mbox{roa-ir}}
\def\pairprop{\mbox{\em pair-propagation}}
\def\tripleprop{\mbox{\em triple-propagation}}
\def\leftfct{{\mbox{${\cal L}${\em ir}}}}
\def\rightfct{{\mbox{${\cal R}${\em ir}}}}
\def\degcasee{{\mbox{\em de}}}
\def\degcased{{\mbox{\em dd}}}
\def\leftn{{\mbox{\em lr}}}
\def\behonen{{\mbox{\em bp}}}
\def\coinconen{{\mbox{\em cp}}}
\def\betwn{{\mbox{\em bw}}}
\def\coinctwon{{\mbox{\em cr}}}
\def\behtwon{{\mbox{\em br}}}
\def\rightn{{\mbox{\em rr}}}
\def\binmat{{\cal B}}
\def\termat{{\cal T}}
\def\north{{\mbox{\em No}}}
\def\northeast{{\mbox{\em NE}}}
\def\east{{\mbox{\em Ea}}}
\def\southeast{{\mbox{\em SE}}}
\def\south{{\mbox{\em So}}}
\def\southwest{{\mbox{\em SW}}}
\def\west{{\mbox{\em We}}}
\def\northwest{{\mbox{\em NW}}}
\def\equal{{\mbox{\em Eq}}}
\def\alc{{\cal ALC}}
\def\atra{{\cal CYC}_t}
\def\BBR{{\rm I\!R}}
\def\cqfd{\vrule height 1.2ex depth 0ex width 1.2ex}
\begin{document}
\mainmatter              % start of the contributions
\title{Integrating cardinal direction relations and other
       orientation relations in Qualitative
       Spatial Reasoning\thanks{This work was partly
       supported by the EU project ``{\em Cognitive
       Vision systems}" (CogVis), under grant
       {\em CogVis IST 2000-29375}.}$\;$\thanks{A preliminary
	version of this work has appeared in the Proceedings
	of the Eighth International Symposium of Artificial
	Intelligence and Mathematics \cite{Isli04a}.}}
\author{Amar Isli}
\authorrunning{Isli}   % abbreviated author list (for running head)
\institute{Fachbereich Informatik, Universit\"at Hamburg,\\
           Vogt-K\"olln-Strasse 30, D-22527 Hamburg, Germany\\
\email{am99i@yahoo.com}
}

\maketitle              % typeset the title of the contribution

THE WORK IS ABOUT COMBINING, AND HANDLING THE INTERACTION OF,
EXISTING QUALITATIVE SPATIAL LANGUAGES. THE WEAKNESS OF A
QUALITATIVE (SPATIAL) LANGUAGE IS THAT IT CAN MAKE ONLY A
FINITE NUMBER OF KNOWN-IN-ADVANCE DISTINCTIONS. THE STRENGTH
IS THAT, (1) SUCH A LANGUAGE IS COGNITIVELY ADEQUATE, IN THAT
THE DISTINCTIONS IT CAN MAKE CORRESPOND TO THOSE THAT A
SPECIFIC CLASS OF APPLICATIONS DO NEED, AND (2) REASONING
ABOUT KNOWLEDGE EXPRESSED IN THE LANGUAGE COSTS MUCH LESS
THAN REASONING ABOUT KNOWLEDGE EXPRESSED IN THE
(QUALITATIVELY) ABSTRACTED QUANTITATIVE LANGUAGE. YET, MANY
NOWADAYS APPLICATIONS, WHILE HAVING NO NEED OF THE WHOLE
EXPRESSIVENESS OF A PURELY QUANTITATIVE LANGUAGE, ARE NOT
FULLY SATISFIED BY JUST ONE SINGLE QUALITATIVE LANGUAGE.
THE SOLUTION PROPOSED BY THE WORK IS THE USE OF THE
INTEGRATION OF EXISTING QUALITATIVE LANGUAGES, SO THAT
THE INTEGRATED QUALITATIVE LANGUAGES COMPENSATE EACH OTHER'S
DEFICIENCIES. THE MOST IMPORTANT ISSUE IS TO PROVIDE
EFFECTIVE PROCEDURES FOR THE HANDLING OF THE INTERACTION
BETWEEN THE INTEGRATED QUALITATIVE LANGUAGES. THE PAPER
PROVIDES SUCH A PROCEDURE, AND SHOWS THAT IT TERMINATES AND
DETECTS INCONSISTENCIES WHICH CANNOT BE DETECTED BY REASONING
SEPARATELY ON EACH OF THE PROPJECTIONS OF THE KNOWLEDGE ONTO
THE INTEGRATED QUALITATIVE LANGUAGES.  ONE POTENTIAL, HIGHLY
21ST-CENTURY, APPLICATION OF THE WORK IS THE PROCESSING OF
MULTI-SATELLITE KNOWLEDGE, EACH SATELLITE SENDING ITS
CONTRIBUTION TO THE GLOBAL KNOWLEDGE IN ITS OWN LANGUAGE. THE
IMPORTANCE OF PROCESSING SUCH HETEROGENOUS, MULTI-SOURCE
KNOWLEDGE IS EASILY SEEN BY IMAGINING WHAT COULD HAVE BEEN
THE SAVING IN HUMAN LIFES IN THE STILL-ONGOING DISASTROUS WAR
IN IRAQ, HAD THE INVOLVED, ``HIGHLY'' OBJECTIVE, INTELLIGENCE
AGENCIES OBJECTIVELY GONE THROUGH THE PROCESSING OF THE
INTERACTION BETWEEN THE DIFFERENT PIECES OF THE MULTI-SOURCE
GATHERED KNOWLEDGE ON WMDs (Weapons of Mass Destruction).

THE VERY FIRST VERSION OF THE WORK,
WHICH I THOUGHT, AND STILL THINK, WAS A HUGE NOVELTY TO GIS\footnote{Geographic
Information Systems.}, I SUBMITTED IT, IN 2001, TO A CONFERENCE WHOSE MAIN
TOPIC IS GIS, THE COSIT CONFERENCE.\footnote{COSIT is an (international)
COnference on Spatial Information Theory.}

- The actual paper is the version of the work exactly as rejected at the special
volume of the journal of the Annals of Mathematics and Artificial Intelligence,
AMAI, dedicated to the 2004 symposium on Artificial Intelligence and
Mathematics, AIM'2004.

- The reviews of the AMAI journal are added after the references, for potential
people interested in objectivity of journals' reviewing processes.

- The reviews of the very first version of the work, as rejected at the
COSIT'2001 conference, are also added after the references, for potential people
interested in objectivity of conferences' reviewing processes.
\begin{abstract}
Integrating different knowledge representation languages is one of the
main topics in Qualitative Spatial Reasoning (QSR). Existing languages
are generally incomparable in terms of expressive
power; as such, their integration compensates each other's
representational deficiencies, and is seen by real applications, such
as Geographic Information Systems (GIS), or robot navigation, as an
answer to the well-known poverty conjecture of qualitative languages
in general, and of QSR languages in particular. Knowledge expressed in
such an integrating language decomposes then into parts, or components,
each expressed in one of the integrated languages. Reasoning internally
within each component of such knowledge involves only the language the
component is expressed in, which is not new. The challenging question
is to come with methods for the interaction of the different components
of such knowledge. With these considerations in mind, we propose a
calculus, $\combalg$, integrating two calculi well-known in QSR:
Frank's projection-based cardinal direction calculus,
$\cdalg$, and a coarser version, $\roalg$, of Freksa's relative
orientation calculus. An original constraint propagation procedure,
$\pcsfc$, for $\combalg$-CSPs is presented, which aims at (1)
achieving \underline{p}ath \underline{c}onsistency ($\mbox{\em Pc}$)
for the $\cdalg$ projection; (2) achieving \underline{s}trong
$\underline{4}$-\underline{c}onsistency ($\mbox{\em S4c}$) for the
$\roalg$ projection; and (3) more ($\mbox{\em +}$) ---the ``+"
consists of the implementation of the interaction between the two
integrated calculi. Dealing with the first two points is not new, and
involves mainly the $\cdalg$ composition table and the $\roalg$
composition table, which can be found in, or derived from, the literature. The
originality of the propagation algorithm comes from the last point.
Two tables, one for each of the two directions $\cdatoroa$
and $\roatocda$, capturing the interaction between the two
kinds of knowledge, are defined, and used by the algorithm. The
importance of taking into account the
interaction is shown with a real example providing an inconsistent
knowledge base, whose inconsistency (a) cannot be detected by
reasoning separately about each of the two components of the
knowledge, just because, taken separately, each is consistent, but
(b) is detected by the proposed algorithm, thanks to the
interaction knowledge propagated from each of the two compnents to
the other.
\end{abstract}
\newtheorem{thr}{Theorem}
\newtheorem{df}{Definition}
\newtheorem{ex}{Example}
\section{Introduction}\label{sect1}
Reasoning about orientation has been, for more than a decade now,
one of the main aspects focussed on in Qualitative Spatial
Reasoning (QSR). A possible explanation stems from the large
number of real applications in need for a qualitative formalism for
representing and reasoning about orientation; among these, we
have Geographic Information Systems (GIS), and robot navigation.
The reader is referred to \cite{Cohn97b} for a survey article on
the different representation techniques, and the different aspects
dealt with, in QSR.

Two important, and widely known, calculi for the representation
and processing of orientation are the projection-based calculus of
cardinal directions, $\cdalg$, in
\cite{Frank92b}, and the relative orientation calculus
in \cite{Freksa92b}. The former uses a global,
west-east/south-north reference frame, and represents knowledge as
binary relations on (pairs of) 2D points. The latter allows for
the representation of relative knowledge as ternary relations on
(triples of) 2D points. Both kinds of knowledge are of particular
importance, especially in large-scale GIS for the former, and in robot
navigation for the latter. An example on high-level satellite-like
surveillance of a geographic area will illustrate that the
integration of the two calculi is much better suited for large-scale GIS
reasoning, than $\cdalg$ alone.

Research in constraint-based QSR has reached a point where the need
for combining, on the one hand, different kinds of existing relations,
such as, in the present work, binary relative orientation relations
based on a global frame of reference (pseudo ternary relations)
\cite{Frank92b} and (purely) ternary relative orientation
relations \cite{Freksa92b}, and, on the other hand, different levels of
local consistency, such as, also in the present work, path consistency
and strong $4$-consistency, is necessary in order to face the
increasing and often challenging demand coming from real applications.

The aim of this work is to look at the importance of integrating the two
orientation calculi mentioned above. Considered separately, the
projection-based calculus in \cite{Frank92b}, $\cdalg$, represents knowledge such as
``Hamburg is north-west of Berlin", whereas the relative orientation
calculus in \cite{Freksa92b} represents knowledge such as ``You
see the main train station on your left when you walk down to the
cinema from the university". We propose a calculus, $\combalg$,
integrating $\cdalg$ and a coarser version, $\roalg$, of the
calculus in \cite{Freksa92b}. $\combalg$ allows for more expressiveness than each of the
integrated calculi, and represents, within the same base, knowledge such
as the one in the following example.
\begin{ex}\label{example1}
\begin{figure}
\centerline{\epsffile{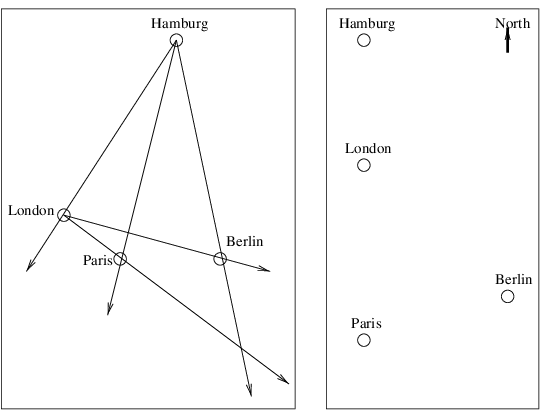}}
\caption{A model for the $\roalg$ component (left), and a model for the
  $\cdalg$ component (right), of the knowledge in Example \ref{example1}.}\label{figure-example1}
\end{figure}
Consider the following knowledge on four cities, Berlin, Hamburg,
London and Paris:
\begin{enumerate}
  \item viewed from Hamburg, Berlin is to the left of Paris, Paris
    is to the left of London, and Berlin is to the left of London;
  \item viewed from London, Berlin is to the left of Paris;
  \item Hamburg is to the north of Paris, and north-west of Berlin; and
  \item Paris is to the south of London.
\end{enumerate}
The first two sentences express the $\roalg$ component of the
knowledge (relative orientation relations on triples of the four
cities), whereas the other two express the $\cdalg$ component of the
knowledge (cardinal direction relations on pairs of the four cities).\footnote{Two
cardinal direction calculi, to be explained later, are known in the literature
\cite{Frank92b}: a cone-shaped and a projection-based (see
illustration in Figure \ref{sectors}). We assume in this example the latter.}
Considered separately, each of the two components is consistent, in
the sense that one can find an assignment of physical locations to
the cities that satisfies all the constraints of the component ---see
the illustration in Figure \ref{figure-example1}.
However, considered globally, the knowledge is clearly inconsistent (from
``viewed from Hamburg, Paris is to the left of London'', we infer that
Hamburg, London and Paris are not collinear -they form a triangle-,
whereas from the conjunction ``Hamburg is to the north of Paris'' and
``Paris is to the south of London'', we infer that Hamburg, London and
Paris are collinear).
\end{ex}
Example \ref{example1} clearly shows that reasoning about combined
knowledge consisting of an $\roalg$ component and a $\cdalg$ component, 
e.g., checking its consistency, does not reduce to a matter
of reasoning about each component separately ---reasoning separately
about each component in the case of Example \ref{example1} shows two
components that are both consistent, whereas the conjunction of the
knowledge in the two components is inconsistent. As a consequence, the
interaction between the two kinds of knowledge has to be handled.
With this in mind, we propose a constraint propagation procedure,
$\pcsfc$, for $\combalg$-CSPs, which aims at:
\begin{enumerate}
  \item achieving \underline{p}ath \underline{c}onsistency
    ($\mbox{\em Pc}$) for the $\cdalg$ projection;
  \item achieving \underline{s}trong
    $\underline{4}$-\underline{c}onsistency ($\mbox{\em S4c}$) for the
    $\roalg$ projection; and
  \item more ($\mbox{\em +}$).
\end{enumerate}
The procedure does more than just achieving path consistency for the $\cdalg$ projection,
and strong $4$-consistency for the $\roalg$ projection.  It implements 
as well the interaction between the two integrated calculi. For this
purpose:
\begin{enumerate}
  \item The procedure makes use, on the one hand, of an augmented composition table of the
    $\cdalg$ calculus:
    \begin{enumerate}
      \item the table records, for each pair $(r,s)$ of $\cdalg$ atoms, the standard
        composition, $r\circ s$, of $r$ and $s$, which is not new, and can be found in the
        literature \cite{Frank92b,Ligozat98a}; and
      \item more importantly, the table records the $\cdatoroa$ interaction, by providing,
        for each pair $(r,s)$ of $\cdalg$ atoms, the most specific $\roalg$ relation,
        $r\otimes s$, such that, for all $x,y,z$, the conjunction $r(x,y)\wedge s(y,z)$
        logically implies $(r\otimes s)(x,y,z)$.
    \end{enumerate}
  \item On the other hand, the procedure makes use of a table for the $\roatocda$
	interaction, providing, for each $\roalg$ atom $t$, the $\cdalg$ constraints it
	imposes on the different pairs of its three arguments.
\end{enumerate}
The procedure is, to the best of our knowledge, original.

The rest of the paper is organised as follows. Section \ref{csps} provides
some background on constraint satisfaction problems (CSPs), on
constraint matrices and on relation algebras. Section \ref{existingcalculi} presents
a quick overview of the cardinal direction calculi in \cite{Frank92b}, and
of the relative orientation calculus in \cite{Freksa92b}. Section \ref{newcalculus} defines a
relative orientation calculus, $\roalg$, which is a caorser version of the one in
\cite{Freksa92b}. Reasoning in the integrating language of $\cdalg$ relations and
$\roalg$ relations is dealt with in detail in Section \ref{reasoning}; in
particular, the section presents the $\cdatoroa$ and the $\roatocda$ interaction tables,
as well as the constraint propagation algorithm $\pcsfc$, both alluded
to above. Section \ref{discussion} provides a short discussion 
relating the work to current research on spatio-temporalising the 
well-known $\alcd$ family of description logics (DLs) with a concrete
domain \cite{Baader91a}: the discussion shows that if two (spatial)
ontologies operate on the same universe of objects (in this work, the
universe of 2D points), while using different languages for their
knowledge representation, then integrating the two ontologies needs an 
inference mechanism for the interaction of the two languages, so that, 
given knowledge expressed in the integrating ontology, consisting of
two components (one for each of the integrated ontologies), each of
the two components can infer knowledge from the other.
Section \ref{summary} summarises the work.
\section{Constraint satisfaction problems}\label{csps}
A constraint satisfaction problem (CSP) of order $n$ consists of:
\begin{enumerate}
  \item a finite set of $n$ variables, $x_1,\ldots ,x_n$;
  \item a set $U$ (called the universe of the problem); and
  \item a set of constraints on values from $U$ which may be assigned to the
    variables.
\end{enumerate}
An $m$-ary constraint is of the form $R(x_{i_1},\cdots ,x_{i_m})$, and asserts
that the $m$-tuple of values assigned to the variables $x_{i_1},\cdots ,x_{i_m}$
must lie in the $m$-ary relation $R$ (an $m$-ary relation over the
universe $U$ is any subset of $U^m$). An $m$-ary CSP is one of which the
constraints are $m$-ary constraints. We will be concerned exclusively
with binary CSPs and ternary CSPs.
 
For any two binary relations $R$ and $S$, $R\cap S$ is the
intersection of $R$ and $S$,
$R\cup S$ is the union of $R$ and $S$, $R\circ S$ is the composition of $R$ and $S$, 
and $R^\smile$ is the converse of $R$; these are defined as follows:
\begin{center}
$
\begin{array}{lll}
R\cap S      &=&\{(a,b):(a,b)\in R\mbox{ and }(a,b)\in S\},\\
R\cup S      &=&\{(a,b):(a,b)\in R\mbox{ or }(a,b)\in S\},\\
R\circ S   &=&\{(a,b):\mbox{for some }c,(a,c)\in R\mbox{ and }(c,b)\in S\},\\
R^\smile     &=&\{(a,b):(b,a)\in R\}.
\end{array}
$
\end{center}
Three special binary relations over a universe $U$ are the empty
relation $\emptyset$ which contains
no pairs at all, the identity relation ${\cal I}_U^b=\{(a,a):a\in U\}$, and the universal
relation $\top _U^b=U\times U$.

Composition and converse for binary relations were introduced by De Morgan
\cite{DeMorgan1864a,DeMorgan66a}. In \cite{Isli00b}, the authors extended the two operations to ternary relations;
furthermore, they introduced for ternary relations the operation of rotation,
which is not needed for binary relations.
For any two ternary relations $R$ and $S$, $R\cap S$ is the intersection of $R$ and $S$,
$R\cup S$ is the union of $R$ and $S$, $R\circ S$ is the composition of $R$ and $S$, 
$R^\smile$ is the converse of $R$, and $R^\frown$ is the rotation of $R$; these are
defined as follows:
\begin{center}
$
\begin{array}{lll}
R\cap S      &=&\{(a,b,c):(a,b,c)\in R\mbox{ and }(a,b,c)\in S\},\\
R\cup S      &=&\{(a,b,c):(a,b,c)\in R\mbox{ or }(a,b,c)\in S\},\\
R\circ S     &=&\{(a,b,c):\mbox{for some }d,(a,b,d)\in R\mbox{ and }(a,d,c)\in S\},\\
R^\smile     &=&\{(a,b,c):(a,c,b)\in R\},\\
R^\frown     &=&\{(a,b,c):(c,a,b)\in R\}.
\end{array}
$
\end{center}
Three special ternary relations over a universe $U$ are the empty relation
$\emptyset$ which contains no triples at all, the identity relation
${\cal I}_U^t=\{(a,a,a):a\in U\}$, and the universal relation
$\top _U^t=U\times U\times U$.
\subsection{Constraint matrices}
A binary constraint matrix of order $n$ over $U$ is an $n\times n$-matrix, say $\binmat$,
of binary relations over $U$ verifying the following:
\begin{center}
$
\begin{array}{ll}
(\forall i\leq n)(\binmat _{ii}\subseteq {\cal I}_U^b)   &\mbox{(the diagonal property)},\\
(\forall i,j\leq n)(\binmat _{ij}=(\binmat _{ji})^\smile )&\mbox{(the converse property)}.
\end{array}
$
\end{center}
A binary CSP $P$ of order $n$ over a universe $U$ can be associated with the
following binary constraint matrix, denoted $\binmat ^P$:
\begin{enumerate}
  \item Initialise all entries to the universal relation:
    $(\forall i,j\leq n)((\binmat ^P)_{ij}\leftarrow \top _U^b)$
  \item Initialise the diagonal elements to the identity relation:\\
    $(\forall i\leq n)((\binmat ^P)_{ii}\leftarrow {\cal I}_U^b)$
  \item For all pairs $(x_i,x_j)$ of variables on which a
    constraint $(x_i,x_j)\in R$ is specified:
    $(\binmat ^P)_{ij}\leftarrow (\binmat ^P)_{ij}\cap R,(\binmat ^P)_{ji}\leftarrow ((\binmat ^P)_{ij})^\smile$.
\end{enumerate}
A ternary constraint matrix of order $n$ over $U$ is an
$n\times n\times n$-matrix, say $\termat$, of ternary relations over $U$ verifying the
following:
\begin{center}
$
\begin{array}{ll}
(\forall i\leq n)(\termat _{iii}\subseteq {\cal I}_U^t)           &\mbox{(the identity property)},\\
(\forall i,j,k\leq n)(\termat _{ijk}=(\termat _{ikj})^\smile )           &\mbox{(the converse property)},\\
(\forall i,j,k\leq n)(\termat _{ijk}=(\termat _{kij})^\frown )           &\mbox{(the rotation property)}.
\end{array}
$
\end{center}
A ternary CSP $P$ of order $n$ over a universe $U$ can be associated with the following
ternary constraint matrix, denoted $\termat ^P$:
\begin{enumerate}
  \item Initialise all entries to the universal relation:\\
    $(\forall i,j,k\leq n)((\termat ^P)_{ijk}\leftarrow \top _U^t)$
  \item Initialise the diagonal elements to the identity relation:\\
    $(\forall i\leq n)((\termat ^P)_{iii}\leftarrow {\cal I}_U^t)$
  \item For all triples $(x_i,x_j,x_k)$ of variables on which
    a constraint $(x_i,x_j,x_k)\in R$ is specified:\\
    $
    \begin{array}{ll}
    (\termat ^P)_{ijk}\leftarrow (\termat ^P)_{ijk}\cap R,     &(\termat ^P)_{ikj}\leftarrow ((\termat ^P)_{ijk})^\smile ,\\
    (\termat ^P)_{jki}\leftarrow ((\termat ^P)_{ijk})^\frown , &(\termat ^P)_{jik}\leftarrow ((\termat ^P)_{jki})^\smile ,\\
    (\termat ^P)_{kij}\leftarrow ((\termat ^P)_{jki})^\frown , &(\termat ^P)_{kji}\leftarrow ((\termat ^P)_{kij})^\smile.
    \end{array}
    $
\end{enumerate}
We make the assumption that, unless explicitly specified otherwise, a CSP is
given as a constraint matrix.
\subsection{Strong $k$-consistency, refinement}
Let $P$ be a CSP of order $n$, $V$ its set of variables and $U$ its universe.
An instantiation of $P$ is any $n$-tuple $(a_1,a_2,\ldots ,a_n)$ of $U^n$,
representing an assignment of a value to each variable.  A consistent instantiation
is an instantiation $(a_1,a_2,\ldots ,a_n)$ which is a solution:
\begin{enumerate}
  \item[$\bullet$] If $P$ is a binary CSP: $(\forall i,j\leq n)((a_i,a_j)\in (\binmat ^P)_{ij})$
  \item[$\bullet$] If $P$ is a ternary CSP: $(\forall i,j,k\leq n)((a_i,a_j,a_k)\in (\termat ^P)_{ijk})$
\end{enumerate}
$P$ is consistent if it has at least one solution; it is inconsistent otherwise. The
consistency problem of $P$ is the problem of verifying whether $P$ is consistent.

Let $V'=\{x_{i_1},\ldots ,x_{i_j}\}$ be a subset of $V$. The sub-CSP of $P$ generated
by $V'$, denoted $P_{|V'}$, is the CSP with $V'$ as the set of variables, and whose constraint
matrix is obtained by projecting the constraint matrix of $P$ onto $V'$:
\begin{enumerate}
  \item[$\bullet$] If $P$ is a binary CSP then: $(\forall k,l\leq j)((\binmat ^{P_{|V'}})_{kl}=(\binmat ^P)_{i_ki_l})$
  \item[$\bullet$] If $P$ is a ternary CSP then: $(\forall k,l,m\leq j)((\termat ^{P_{|V'}})_{klm}=(\termat ^P)_{i_ki_li_m})$
\end{enumerate}
$P$ is $k$-consistent \cite{Freuder78a,Freuder82a} if for any subset $V'$ of $V$
containing $k-1$ variables, and for any variable $X\in V$, every solution to
$P_{|V'}$ can be extended to a solution to $P_{|V'\cup\{X\}}$. $P$ is strongly
$k$-consistent if it is $j$-consistent, for all $j\leq k$.

$1$-consistency, $2$-consistency and $3$-consistency correspond to node-consistency,
arc-consistency and path-consistency, respectively \cite{Mackworth77a,Montanari74a}.
Strong $n$-consistency of $P$ corresponds to what is called global consistency in
\cite{Dechter92a}. Global consistency facilitates the important task of searching
for a solution, which can be done, when the property is met, without backtracking
\cite{Freuder82a}.

A refinement of $P$ is a CSP $P'$ with the same set of variables, and such that
\begin{enumerate}
  \item[$\bullet$] $(\forall i,j)((\binmat ^{P'})_{ij}\subseteq (\binmat ^P)_{ij})$, in the case of binary CSPs.
  \item[$\bullet$] $(\forall i,j,k)((\termat ^{P'})_{ijk}\subseteq (\termat ^P)_{ijk})$, in the case of ternary CSPs.
\end{enumerate}
\subsection{Relation algebras}
The reader is referred to \cite{Tarski41b,Ladkin94a} for the
definition of a binary Relation Algebra (RA), and to \cite{Isli00b} for
the definition of a ternary RA. Of particular interest to this work
are:
\begin{enumerate}
  \item binary RAs of the form
    $\langle {\cal A},\oplus ,\odot,^-,\bot ,\top ,\circ ,^\smile ,{\cal I}\rangle$,
    where ${\cal A}$ is a non empty finite set, and $\circ$ and $^\smile$ are the
    operations of composition and converse, respectively; and
  \item ternary RAs of the form
    $\langle {\cal A},\oplus ,\odot,^-,\bot ,\top ,\circ ,^\smile ,^\frown ,{\cal I}\rangle$,
    where ${\cal A}$ is a non empty finite set, and $\circ$, $^\smile$
    and $^\frown$ are the operations of composition, converse and
    rotation, respectively.
\end{enumerate}
\section{Existing orientation calculi}\label{existingcalculi}
Some background on existing orientation calculi is in order.
\subsection{The cardinal direction calculi in \cite{Frank92b}}
\begin{figure}[t]
\centerline{\epsfxsize=11cm\epsfysize=4.75cm\epsffile{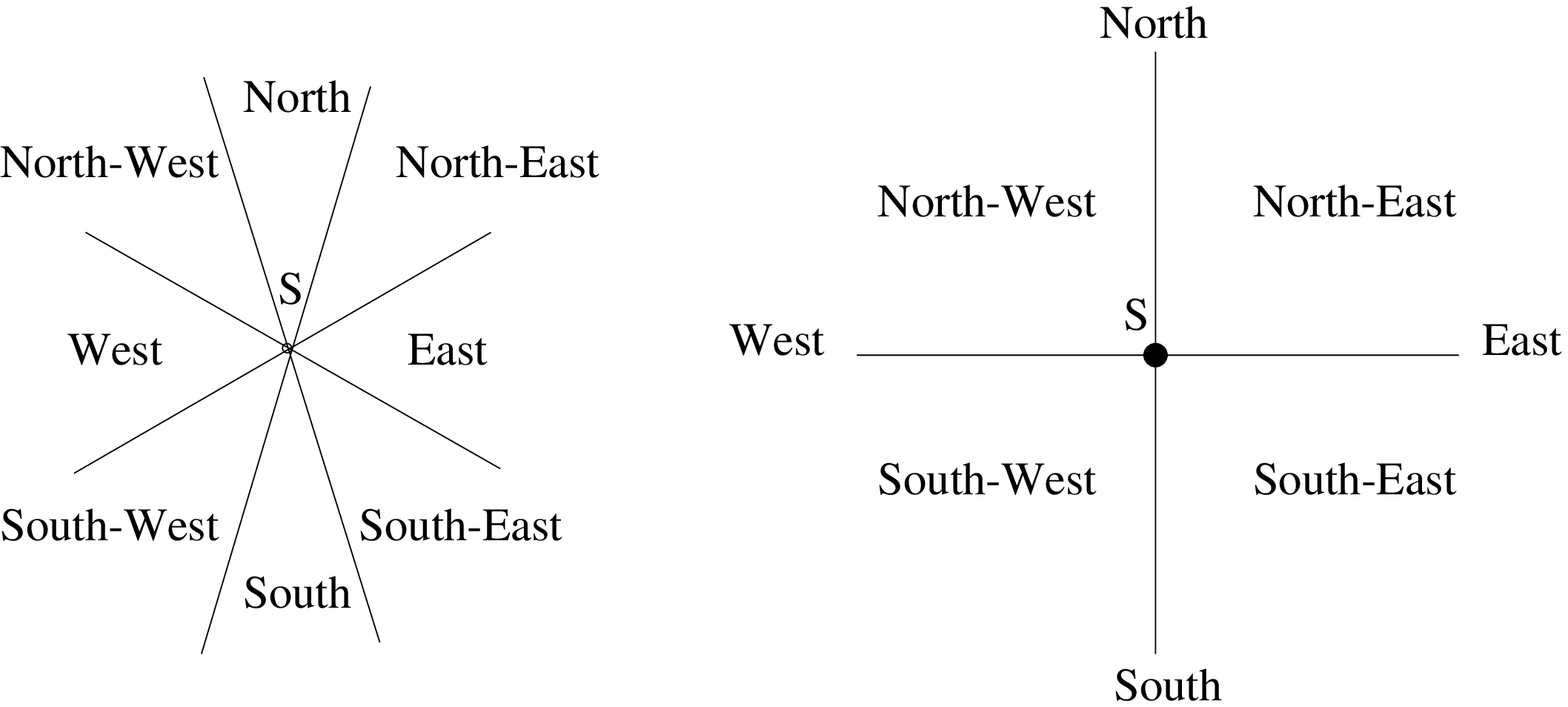}}
\caption{The cone-shaped (left) and projection-based (right) models of
cardinal directions in \cite{Frank92b}.}\label{sectors}
\end{figure}
The models of cardinal directions in 2D developed in \cite{Frank92b} are
illustrated in Figure \ref{sectors}. They use a partition of the plane
into regions determined by lines
passing through a reference object, say $S$. Depending on the region
a point $P$ belongs to, we have $\north (P,S)$, $\northeast (P,S)$, $\east (P,S)$, 
$\southeast (P,S)$, $\south (P,S)$, $\southwest (P,S)$, $\west (P,S)$, $\northwest (P,S)$, or
$\equal (P,S)$, corresponding, respectively, to the position of $P$
relative to $S$ being $\mbox{\em north}$, $\mbox{\em north-east}$,
$\mbox{\em east}$, $\mbox{\em south-east}$, $\mbox{\em south}$,
$\mbox{\em south-west}$, $\mbox{\em west}$, $\mbox{\em north-west}$,
or $\mbox{\em equal}$. Each of the two models can thus be seen as a
binary RA, with nine atoms.
Both use a global, {\em west-east/south-north}, reference frame.
We focus our attention on the projection-based model (Figure \ref{sectors}(right)), which has been
assessed as being cognitively more adequate \cite{Frank92b}
(cognitive adequacy of spatial orientation models is discussed in
\cite{Freksa92b}).
\subsection{The relative orientation calculus in \cite{Freksa92b}}
\begin{figure}[t]
\centerline{\epsfxsize=10.5cm\epsfysize=5.1cm\epsffile{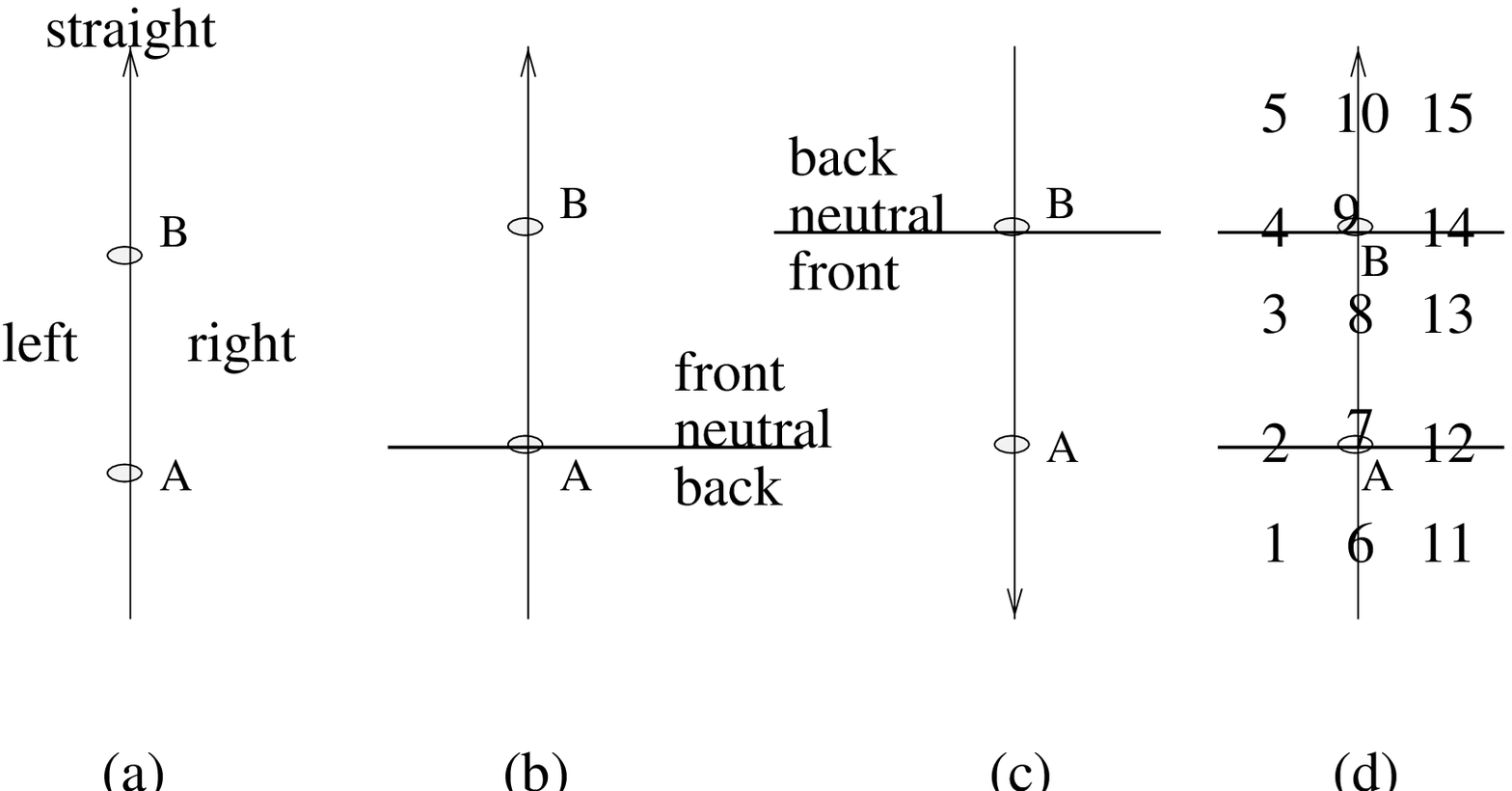}}
\caption{The partition of the universe of 2D positions on which is based
         the relative orientation calculus in \cite{Freksa92b}.}\label{rel-orient}
\end{figure}
A well-known model of relative orientation of 2D points is the
calculus in \cite{Freksa92b}. It is derived from a specific partition, into $15$
regions, of the plane, determined by a parent object, say $A$, and a reference
object, say $B$ (Figure \ref{rel-orient}(d)). The partition is based on the
following:
\begin{enumerate}
  \item the {\it left/straight/right} partition of the plane determined by an observer placed
    at the parent object and looking in the direction of the reference object
    (Figure \ref{rel-orient}(a));
  \item the {\it front/neutral/back} partition of the plane determined by the same observer
    (Figure \ref{rel-orient}(b)); and
  \item the similar {\it front/neutral/back} partition of the plane obtained when we swap the
    roles of the parent object and the reference object (Figure \ref{rel-orient}(c)).
\end{enumerate}
Combining the three partitions (a), (b) and (c) of Figure \ref{rel-orient}
leads to the partition of the universe of 2D positions on which is based the
calculus in \cite{Freksa92b} (Figure \ref{rel-orient}(d)).
\section{A new relative orientation calculus}\label{newcalculus}
The projection-based model of cardinal directions in \cite{Frank92b} uses a global,
west-east/south-north, reference frame; its use and importance in GIS 
are well-known. The calculus in \cite{Freksa92b} is more suited
for the description of a configuration of 2D points (a spatial scene) relative
to one another. Integrating the two kinds of relations
would lead to more expressiveness than allowed by each of the integrated
calculi, so that one would then be able to
represent, within the same base, knowledge such as the one in
the 4-sentence example provided in the introduction.

\begin{figure}
\centerline{\epsfxsize=7cm\epsfysize=3.4cm\epsffile{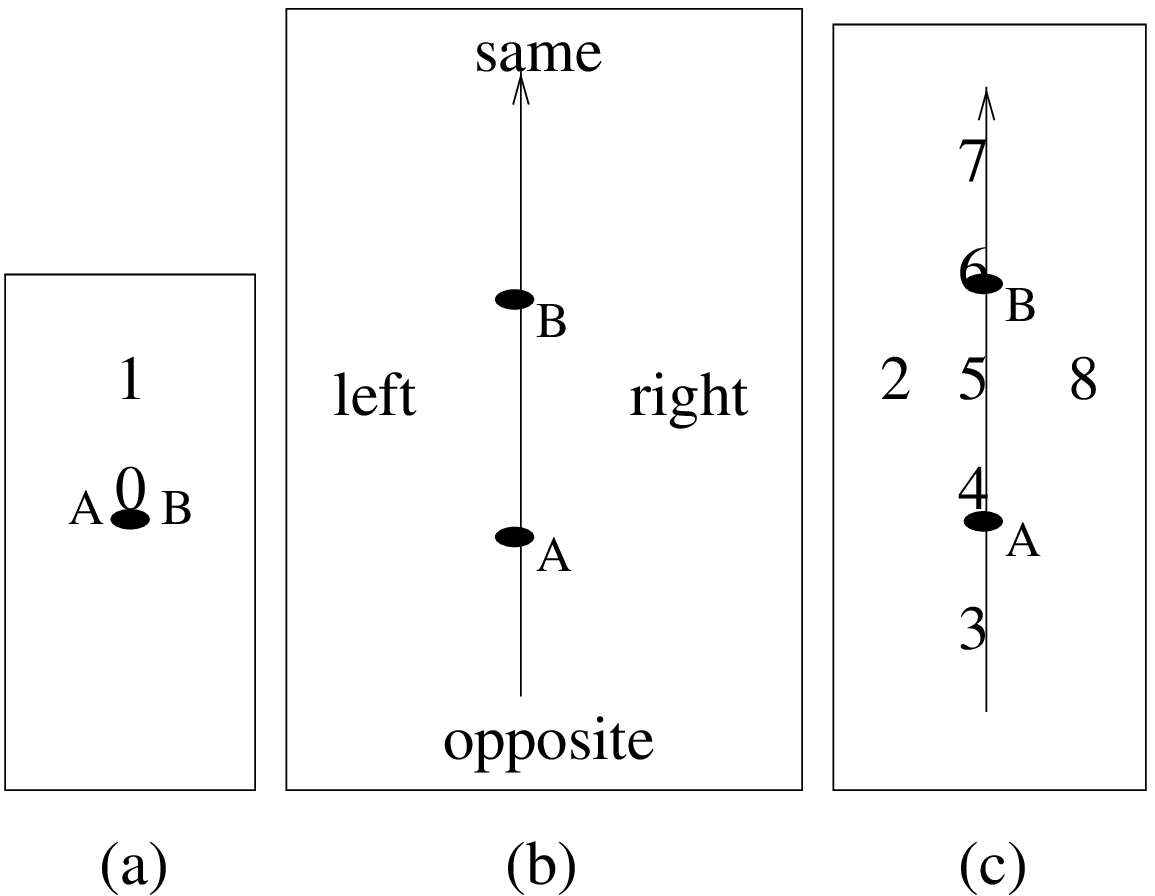}}
\caption{The partition of the universe of 2D positions on which is based
         the $\roalg$ calculus.}\label{rel-orient-n}
\end{figure}

The coarser relative orientation calculus can be obtained from the one in \cite{Freksa92b} by ignoring, in the construction of the partition of the plane
determined by a parent object and a reference object (Figure
\ref{rel-orient}(d)), the two {\it front/neutral/back} partitions
(Figure \ref{rel-orient}(b-c)). In other words, we consider only the
{\it left/straight/right} partition (Figure \ref{rel-orient}(a)) ---we 
also keep the 5-element partitioning of the line joining the parent
object to the reference object.
The final situation is depicted in Figure \ref{rel-orient-n}, where $A$ and $B$
are the parent object and the reference object, respectively:
\begin{enumerate}
  \item Figure \ref{rel-orient-n}(b-c) depicts the general case, corresponding to
    the parent object and the reference object being distinct from
    each other: this
    general-case partition leads to $7$ regions (Figure \ref{rel-orient-n}(c)),
    numbered from $2$ to $8$, corresponding to 7 of the
    nine atoms of the calculus, which we refer to as
    $\leftn$ (to the \underline{\em l}eft of the \underline{\em r}eference object), $\behonen$
    (\underline{\em b}ehind the \underline{\em p}arent object),
    $\coinconen$ (\underline{\em c}oincides with the \underline{\em
      p}arent object), $\betwn$ (\underline{\em b}et\underline{\em w}een the parent object and the reference object),
      $\coinctwon$ (\underline{\em c}oincides with the \underline{\em
      r}eference object), $\behtwon$  (\underline{\em b}ehind the \underline{\em
      r}eference object), and $\rightn$ (to the \underline{\em r}ight of the \underline{\em r}eference object).
  \item Figure \ref{rel-orient-n}(a) illustrates the degenerate case, corresponding 
    to equality of the parent object and the reference object. The two 
    regions, corresponding, respectively, to the primary object coinciding with the
    parent object and the reference object, and to the primary object
    distinct from the parent object and the reference object, are numbered 
    0 and 1. The corresponding atoms of the calculus will
    be referred to as $\degcasee$ (\underline{\em d}egenerate \underline{\em e}qual)
    and $\degcased$ (\underline{\em d}egenerate \underline{\em d}istinct).
\end{enumerate}
From now on, we refer to the calculus in \cite{Frank92b} as $\cdalg$
(Cardinal Direction Algebra),
and to the coarser version of the calculus in \cite{Freksa92b} as $\roalg$ (Relative
Orientation Algebra). A $\cdalg$ (resp. $\roalg$) relation is any
subset of the set of all $\cdalg$ (resp. $\roalg$) atoms. A $\cdalg$ (resp.
$\roalg$) relation is said to be atomic if it contains one single atom (a
singleton set); it is said to be the $\cdalg$ (resp. $\roalg$) universal
relation if it contains all the $\cdalg$ (resp. $\roalg$) atoms. When no
confusion raises, we may omit the brackets in the representation of an
atomic relation.\\
\section{Reasoning about combined
  knowledge of $\cdalg$ relations and $\roalg$ relations}\label{reasoning}
We start now the main part of the paper, i.e., the representation of
knowledge about 2D points as a combined conjunction of:
\begin{enumerate}
  \item $\cdalg$ relations on (pairs of) the objects, on the one hand;
    and
  \item $\roalg$ relations on (triples of) the objects, on the other
    hand.
\end{enumerate}
More importantly, we deal with the issue of reasoning about such a
combined knowledge. We first present for each of the integrated calculi,
$\cdalg$ and $\roalg$:
\begin{enumerate}
  \item tables recording the internal reasoning: the tables of converse
    and composition for $\cdalg$, which can be found in the literature
    \cite{Frank92b,Ligozat98a}; and the tables of converse,
    rotation and composition for $\roalg$, which can be derived from
    the work in \cite{Isli00b}; and
  \item a table for the interaction with the other calculus: a
    $\cdatoroa$ interaction table, recording the $\roalg$ knowledge
    inferred from $\cdalg$ knowledge; and an $\roatocda$ interaction
    table, recording the $\cdalg$ knowledge inferred from $\roalg$
    knowledge.
\end{enumerate}
We then give a quick presentation of what
is already known in the literature: CSPs of $\cdalg$ relations
\cite{Frank92b,Ligozat98a}, and
the way to solve them \cite{Ligozat98a}. Then come the definition of CSPs
of $\roalg$ relations, and a discussion on how to adapt a known
propagation algorithm \cite{Isli00b} to such CSPs. We
finish the section with the presentation of CSPs combining both kinds of knowledge
(CSPs of $\cdalg$ relations and $\roalg$ relations on 2D points): most
importantly, this last part will present in detail the propagation
algorithm $\pcsfc$ we have already alluded to.
\subsection{Reasoning within $\cdalg$ and the $\cdatoroa$ interaction: the tables}
The table in Figure \ref{cda-comp} presents the augmented $\cdalg$ composition
table; for each pair $(r_1,r_2)$ of $\cdalg$ atoms, the
table provides:
\begin{enumerate}
  \item the standard composition, $r_1\circ r_2$, of $r_1$ and
    $r_2$ \cite{Frank92b,Ligozat98a}; and
  \item the most specific $\roalg$ relation $r_1\otimes r_2$ such
    that, for all 2D points $x,y,z$, the conjunction
    $r_1(x,y)\wedge r_2(y,z)$ logically implies $(r_1\otimes
    r_2)(x,y,z)$.
\end{enumerate}
\begin{figure}
\begin{tiny}
\begin{center}
$
\begin{array}{|l||l|l|l|l|l|l|l|l|} \hline
{\frac \circ \otimes}      &\north
                &\south
                     &\east
                          &\west
                               &\northeast
                                    &\northwest
                                         &\southeast
                                              &\southwest  \\  \hline\hline
\north
           &\north
                &[\south ,\north ]
                     &\northeast
                          &\northwest
                               &\northeast
                                    &\northwest
                                         &[\southeast ,\northeast ]
                                              &[\southwest ,\northwest 
                                              ]  \\  \cline{2-9}
           &\behtwon
                &\{\behonen ,\coinconen ,\betwn\}
                     &\rightn
                          &\leftn
                               &\rightn
                                    &\leftn
                                         &\rightn
                                              &\leftn  \\  \hline
\south     &[\south ,\north ]
                &\south
                     &\southeast
                          &\southwest
                               &[\southeast ,\northeast ]
                                    &[\southwest ,\northwest ]
                                         &\southeast
                                              &\southwest  \\  \cline{2-9}
           &\{\behonen ,\coinconen ,\betwn\}
                &\behtwon
                     &\leftn
                          &\rightn
                               &\leftn
                                    &\rightn
                                         &\leftn
                                              &\rightn  \\  \hline
\east      &\northeast
                &\southeast
                     &\east
                          &[\west ,\east ]
                               &\northeast
                                    &[\northwest ,\northeast ]
                                         &\southeast
                                              &[\southwest ,\southeast ]  \\  \cline{2-9}
           &\leftn
                &\rightn
                     &\behtwon
                          &\{\behonen ,\coinconen ,\betwn\}
                               &\leftn
                                    &\leftn
                                         &\rightn
                                              &\rightn  \\  \hline
\west      &\northwest
                &\southwest
                     &[\west ,\east ]
                          &\west
                               &[\northwest ,\northeast ]
                                    &\northwest
                                         &[\southwest ,\southeast ]
                                              &\southwest  \\  \cline{2-9}
           &\rightn
                &\leftn
                     &\{\behonen ,\coinconen ,\betwn\}
                          &\behtwon
                               &\rightn
                                    &\rightn
                                         &\leftn
                                              &\leftn  \\  \hline
\northeast &\northeast
                &[\southeast ,\northeast ]
                     &\northeast
                          &[\northwest ,\northeast ]
                               &\northeast
                                    &[\northwest ,\northeast ]
                                         &[\southeast ,\northeast ]
                                              &?  \\  \cline{2-9}
           &\leftn
                &\rightn
                     &\rightn
                          &\leftn
                               &\{\leftn ,\behtwon ,\rightn\}
                                    &\leftn
                                         &\rightn
                                              &\{\leftn ,\behonen ,\coinconen ,  \\
           &
                &
                     &
                          &
                               &
                                    &
                                         &
                                              &\betwn ,\rightn\}  \\  \hline
\northwest &\northwest
                &[\southwest ,\northwest ]
                     &[\northwest ,\northeast ]
                          &\northwest
                               &[\northwest ,\northeast ]
                                    &\northwest
                                         &?
                                              &[\southwest ,\northwest ]  \\  \cline{2-9}
           &\rightn
                &\leftn
                     &\rightn
                          &\leftn
                               &\rightn
                                    &\{\leftn ,\behtwon ,\rightn\}
                                         &\{\leftn ,\behonen ,\coinconen ,
                                              &\leftn  \\
           &
                &
                     &
                          &
                               &
                                    &
                                         &\betwn ,\rightn\}
                                              &  \\  \hline
\southeast &[\southeast ,\northeast ]
                &\southeast
                     &\southeast
                          &[\southwest ,\northeast ]
                               &[\southeast ,\northeast ]
                                    &?
                                         &\southeast
                                              &[\southwest ,\northeast ]  \\  \cline{2-9}
           &\leftn
                &\rightn
                     &\leftn
                          &\rightn
                               &\leftn
                                    &\{\leftn ,\behonen ,\coinconen ,
                                         &\{\leftn ,\behtwon ,\rightn\}
                                              &\rightn  \\
           &
                &
                     &
                          &
                               &
                                    &\betwn ,\rightn\}
                                         &
                                              &  \\  \hline
\southwest &[\southwest ,\northwest ]
                &\southwest
                     &[\southwest ,\southeast ]
                          &\southwest
                               &?
                                    &[\southwest ,\northwest ]
                                         &[\southwest ,\southeast ]
                                              &\southwest  \\  \cline{2-9}
           &\rightn
                &\leftn
                     &\leftn
                          &\rightn
                               &\{\leftn ,\behonen ,\coinconen ,
                                    &\rightn
                                         &\leftn
                                              &\{\leftn ,\behtwon ,\rightn\}  \\
           &
                &
                     &
                          &
                               &\betwn ,\rightn\}
                                    &
                                         &
                                              &  \\  \hline
\end{array}
$
\end{center}
\end{tiny}
\caption{The augmented composition table of the projection-based cardinal direction
  calculus in \cite{Frank92b}: for each pair $(r_1,r_2)$ of $\cdalg$ atoms,
  the table provides the composition, $r_1\circ r_2$, of $r_1$ and
  $r_2$, as well as the most specific $\roalg$ relation
  $r_1\otimes r_2$ such that, for all 2D points $x,y,z$, the
  conjunction $r_1(x,y)\wedge r_2(y,z)$ logically implies
  $(r_1\otimes r_2)(x,y,z)$. The question mark symbol ? represents the 
  $\cdalg$ universal relation $\{\north ,\northwest ,\west ,\southwest 
  ,\south ,\southeast ,\east ,\northeast ,\equal\}$.}\label{cda-comp}
\end{figure}
The operation $\circ$ is just the normal composition: it is internal to $\cdalg$, in the sense that it takes
as input two $\cdalg$ atoms, and outputs a $\cdalg$ relation. The operation $\otimes$, however, is not
internal to $\cdalg$, in the sense that it takes as input two $\cdalg$
atoms, but outputs an $\roalg$ relation; $\otimes$ captures the
interaction between $\cdalg$ knowledge and $\roalg$ knowledge, in the
direction $\cdatoroa$, by inferring $\roalg$ knowledge from given
$\cdalg$ knowledge. As an example for the new operation $\otimes$, from
$$\southeast (Berlin,London)\wedge\north (London,Paris),$$
saying that Berlin is south-east of
London, and that London is north of Paris, we infer the $\roalg$
relation $\leftn$ on the triple $(Berlin,London,Paris)$:
$$\leftn (Berlin,London,Paris),$$
saying that, viewed from Berlin, Paris is to the left of
London. As another example, from
$$\north (Paris,Rome)\wedge\south (Rome,London),$$
the most specific $\roalg$ relation we can infer  on the triple 
$(Paris,Rome,London)$ is $\{\behonen ,\coinconen ,\betwn\}$:
$$\{\behonen ,\coinconen ,\betwn\}(Paris,Rome,London).$$
The reader is referred to \cite{Frank92b,Ligozat98a} for the
$\cdalg$ converse table, providing the converse $r^\smile$ for each $\cdalg$ atom $r$.
\subsection{Reasoning within $\roalg$ and the $\roatocda$ interaction: the tables}
Figure \ref{roa-comp-one} provides for each of the $\roalg$ atoms, say $t$, the converse $t^\smile$
  and the rotation $t^\frown$ of $t$. Figure \ref{roa-comp-two} provides the
  $\roalg$ composition tables, which are computed in the following way.
Given four 2D points $x,y,z,w$ and two $\roalg$ atoms $t_1$
and $t_2$, the conjunction $t_1(x,y,z)\wedge t_2(x,z,w)$ is
inconsistent if the most specific relation $b_1(x,z)$, one can infer
from $t_1(x,y,z)$ on the pair $(x,z)$, is different from the most
specific relation $b_2(x,z)$, one can infer from $t_2(x,z,w)$ on the
same pair $(x,z)$. The $\roalg$ composition splits therefore into two
composition tables, one for each of the following two
cases:\footnote{A similar way of splitting the composition into more
than one table has been followed for the ternary RA, $\atra$,
presented in \cite{Isli00b}.}
\begin{enumerate}
  \item Case 1: $x=z$ (i.e., each of $b_1$ and $b_2$ is the relation $=$). This corresponds to $t_1\in\{\degcasee ,\coinconen\}$
    and $t_2\in\{\degcasee ,\degcased\}$.
  \item Case 2: $x\not = z$ (i.e., each of $b_1$ and $b_2$ is the
    relation $\not =$). This corresponds to $t_1\in\{\degcased ,
    \leftn ,\behonen ,\coinconen ,\betwn ,\behtwon ,\rightn\}$
    and $t_2\in\{\leftn ,\behonen ,\coinconen ,\betwn ,\coinctwon ,\behtwon ,\rightn\}$.
\end{enumerate}

The $\cdalg$ knowledge one can
  infer from $\roalg$ relations is presented in the table of Figure
  \ref{roa-comp-three}, which makes use of the following two
  functions, $\leftfct$ and $\rightfct$:\\
\begin{scriptsize}
\begin{center}
$
\leftfct (r)=
  \left\{
                   \begin{array}{ll}
                         \{\southeast ,\east ,\northeast\}
                               &\mbox{ if }r=\south ,  \\
                         \{\southeast ,\east ,\northeast ,\north ,\northwest\}
                               &\mbox{ if }r=\southeast ,  \\
                         \{\northeast ,\north ,\northwest\}
                               &\mbox{ if }r=\east ,  \\
                         \{\northeast ,\north ,\northwest ,\west ,\southwest\}
                               &\mbox{ if }r=\northeast ,  \\
                         \{\northwest ,\west ,\southwest\}
                               &\mbox{ if }r=\north ,  \\
                         \{\northwest ,\west ,\southwest ,\south ,\southeast\}
                               &\mbox{ if }r=\northwest ,  \\
                         \{\southwest ,\south ,\southeast\}
                               &\mbox{ if }r=\west ,  \\
                         \{\southwest ,\south ,\southeast ,\east ,\northeast\}
                               &\mbox{ if }r=\southwest .  \\
                   \end{array}
  \right.
$\hskip 0.1cm
$
\rightfct (r)=
  \left\{
                   \begin{array}{ll}
                         \{\northwest ,\west ,\southwest\}
                               &\mbox{ if }r=\south ,  \\
                         \{\northwest ,\west ,\southwest ,\south ,\southeast\}
                               &\mbox{ if }r=\southeast ,  \\
                         \{\southwest ,\south ,\southeast\}
                               &\mbox{ if }r=\east ,  \\
                         \{\southwest ,\south ,\southeast ,\east ,\northeast\}
                               &\mbox{ if }r=\northeast ,  \\
                         \{\southeast ,\east ,\northeast\}
                               &\mbox{ if }r=\north ,  \\
                         \{\southeast ,\east ,\northeast ,\north ,\southwest\}
                               &\mbox{ if }r=\northwest ,  \\
                         \{\northeast ,\north ,\northwest\}
                               &\mbox{ if }r=\west ,  \\
                         \{\northeast ,\north ,\northwest ,\west ,\southwest\}
                               &\mbox{ if }r=\southwest .  \\
                   \end{array}
  \right.
$
\end{center}
\end{scriptsize}
The function $\leftfct$ (${\cal L}$eft inferred relation) provides for
its argument, say $r$ (a $\cdalg$ atom), the most specific
$\cdalg$ relation $R$ such that for all $x,y,z$, the conjunction
$r(x,y)\wedge\leftn (x,y,z)$ logically implies $R(x,z)$. For instance, 
if $r$ is $\south$ then $R=\leftfct (\south )=\{\southeast ,\east ,\northeast\}$
---from $\south (Paris,London)$ and $\leftn (Paris,London,Madrid)$, we get\\
$\{\southeast ,\east ,\northeast\}(Paris,Madrid)$. As another
example, if
$r$ is $\southeast$ then $R=\leftfct (\southeast )=\{\southeast ,\east
,\northeast ,\north ,\northwest\}$ ---see the illustration of Figure
\ref{eg-bhp}: from $\southeast (Berlin,Hamburg)$ and $\leftn (Berlin,Hamburg,Paris)$,
we get $\{\southeast ,\east ,\northeast ,\north ,\northwest\}(Berlin,Paris)$.
The function
$\rightfct$ (${\cal R}$ight inferred relation) is defined in a similar 
way, with $\leftn$ replaced with $\rightn$.
\begin{figure}[t]
\begin{footnotesize}
\begin{center}
$
\begin{array}{|c||c|c|c|c|c|c|c|c|c|}  \hline
\mbox{Region }R    &0&1&2&3&4&5&6&7&8  \\  \hline\hline
\mbox{Atom }t      &\degcasee
                     &\degcased
                       &\leftn
                         &\behonen
                           &\coinconen
                             &\betwn
                               &\coinctwon
                                 &\behtwon
                                   &\rightn  \\  \hline
t^\smile           &\degcasee
                     &\coinconen
                       &\rightn
                         &\behonen
                           &\degcased
                             &\behtwon
                               &\coinctwon
                                 &\betwn
                                   &\leftn  \\  \hline
t^\frown           &\degcasee
                     &\coinconen
                       &\leftn
                         &\betwn
                           &\coinctwon
                             &\behtwon
                               &\degcased
                                 &\behonen
                                   &\rightn  \\  \hline
\end{array}
$
\end{center}
\end{footnotesize}
\caption{For each of the nine regions $0,\ldots ,8$ in Figure
\ref{rel-orient-n}(a-c), the corresponding $\roalg$ atom $t$, as well as the
converse $t^\smile$ and the rotation $t^\frown$ of $t$.}\label{roa-comp-one}
\end{figure}
\begin{figure}[t]
\begin{footnotesize}
\begin{center}
$
\begin{array}{|l||l|l|} \hline
\circ      &\degcasee
                &\degcased  \\  \hline\hline
\degcasee
           &\degcasee
                &\degcased  \\  \hline%  \cline{2-3}
\coinconen
           &\coinconen
                &\{\leftn ,\behonen ,\betwn ,\coinctwon ,\behtwon
                ,\rightn\}  \\  \hline%  \cline{2-3}
\end{array}
$\vskip 0.3cm
$
\begin{array}{|l||l|l|l|l|l|l|l|} \hline
\circ      &\leftn
                &\behonen
                     &\coinconen
                          &\betwn
                               &\coinctwon
                                    &\behtwon
                                         &\rightn  \\  \hline\hline
\degcased
           &\degcased
                &\degcased
                     &\degcasee
                          &\degcased
                               &\degcased
                                    &\degcased
                                         &\degcased  \\  \hline
\leftn     &\{\leftn ,\behonen ,\rightn\}
                &\rightn
                     &\coinconen
                          &\leftn
                               &\leftn
                                    &\leftn
                                         &\{\leftn ,\betwn ,\coinctwon ,\behtwon ,\rightn\}\\  \hline
\behonen   &\rightn
                &\{\betwn ,\coinctwon ,\behtwon\}
                     &\coinconen
                          &\behonen
                               &\behonen
                                    &\behonen
                                         &\leftn  \\  \hline
\betwn     &\leftn
                &\behonen
                     &\coinconen
                          &\betwn
                               &\betwn
                                    &\{\betwn ,\coinctwon ,\behtwon\}
                                         &\rightn  \\  \hline
\coinctwon &\leftn
                &\behonen
                     &\coinconen
                          &\betwn
                               &\coinctwon
                                    &\behtwon
                                         &\rightn  \\  \hline
\behtwon   &\leftn
                &\behonen
                     &\coinconen
                          &\{\betwn ,\coinctwon ,\behtwon\}
                               &\behtwon
                                    &\behtwon
                                         &\rightn  \\  \hline
\rightn    &\{\leftn ,\betwn ,\coinctwon ,\behtwon ,\rightn\}
                &\leftn
                     &\coinconen
                          &\rightn
                               &\rightn
                                    &\rightn
                                         &\{\leftn ,\behonen
                                         ,\rightn\}  \\  \hline%  \cline{2-8}
\end{array}
$
\end{center}
\end{footnotesize}
\caption{The $\roalg$ composition tables: in each of the two tables,
  the entry at the intersection of a line $\ell$ and a column $c$ is
  the composition, $r_1\circ r_2$, of $r_1$ and $r_2$, where $r_1$ is
  the $\roalg$ atom appearing as the leftmost element of line
  $\ell$ and $r_2$ is the $\roalg$ atom appearing as the
  top element of column $c$.}\label{roa-comp-two}
\end{figure}
\begin{figure}[t]
\begin{footnotesize}
\begin{center}
$
\begin{array}{|l|l|l|l|}  \hline
r
           &\fctroatocdaontw (r,P,i,j,k)
           &\fctroatocdaonth (r,P,i,j,k)
           &\fctroatocdatwth (r,P,i,j,k)  \\  \hline\hline
\degcasee
           &(\binmat ^P)_{ij}\cap\{\equal\}
           &(\binmat ^P)_{ik}\cap\{\equal\}
           &(\binmat ^P)_{jk}\cap\{\equal\}\\  \hline
\degcased
           &(\binmat ^P)_{ij}\cap\{\equal\}
           &(\binmat ^P)_{ik}\cap (\binmat ^P)_{jk}\cap\overline{\{\equal\}}
           &(\binmat ^P)_{ik}\\  \hline
\leftn
           &(\binmat ^P)_{ij}\cap\overline{\{\equal\}}
           &(\binmat ^P)_{ik}\cap\leftfct ((\binmat ^P)_{ij})
           &(\binmat ^P)_{jk}\cap\rightfct ((\binmat ^P)_{ji})\\  \hline
\behonen
           &(\binmat ^P)_{ij}\cap (\binmat ^P)_{ki}\cap (\binmat ^P)_{kj}\cap\overline{\{\equal\}}
           &(\binmat ^P)_{ji}
           &(\binmat ^P)_{ik}\\  \hline
\coinconen
           &(\binmat ^P)_{ij}\cap (\binmat ^P)_{kj}\cap\overline{\{\equal\}}
           &(\binmat ^P)_{ik}\cap\{\equal\}
           &(\binmat ^P)_{ji}\\  \hline
\betwn
           &(\binmat ^P)_{ij}\cap (\binmat ^P)_{ik}\cap (\binmat ^P)_{kj}\cap\overline{\{\equal\}}
           &(\binmat ^P)_{ij}
           &(\binmat ^P)_{ji}\\  \hline
\coinctwon
           &(\binmat ^P)_{ij}\cap (\binmat ^P)_{ik}\cap\overline{\{\equal\}}
           &(\binmat ^P)_{ij}
           &(\binmat ^P)_{jk}\cap\{\equal\}\\  \hline
\behtwon
           &(\binmat ^P)_{ij}\cap (\binmat ^P)_{ik}\cap (\binmat ^P)_{jk}\cap\overline{\{\equal\}}
           &(\binmat ^P)_{ij}
           &(\binmat ^P)_{ij}\\  \hline
\rightn
           &(\binmat ^P)_{ij}\cap\overline{\{\equal\}}
           &(\binmat ^P)_{ik}\cap\rightfct ((\binmat ^P)_{ij})
           &(\binmat ^P)_{jk}\cap\leftfct ((\binmat ^P)_{ji})\\  \hline
\end{array}
$
\end{center}
\end{footnotesize}
\caption{Given a $\combalg$-CSP $P$, the constraints imposed by the $\roalg$ relation
$(\termat ^P)_{ijk}$ on the $\cdalg$ relations on the different pairs of the three
arguments. The table presents the case when $(\termat ^P)_{ijk}$ is
an atomic relation, say $r$; the case when $(\termat ^P)_{ijk}$ is a
disjunctive $\roalg$ relation is explained in the main text.}\label{roa-comp-three}
\end{figure}
\begin{figure}[t]
\centerline{\epsffile{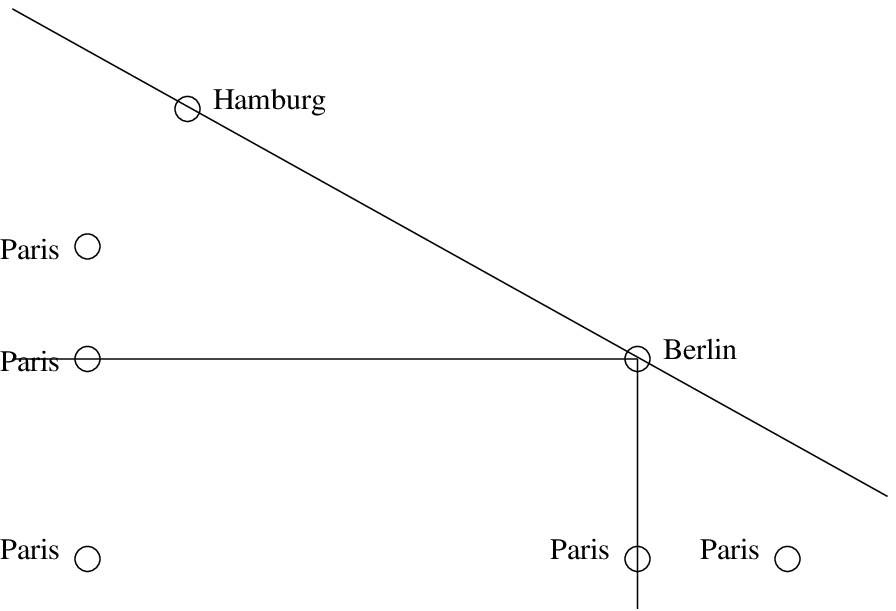}}
\caption{From ``Berlin is south-east of Hamburg" and ``viewed from
  Berlin, Paris is to the left of Hamburg", we infer that ``Berlin is
  south-east, east, north-east, north, or north-west of, Paris".}\label{eg-bhp}
\end{figure}

Given a $\combalg$-CSP $P$, the table in Figure
\ref{roa-comp-three} illustrates how the $\roalg$ constraint $(\termat
^P)_{ijk}$ on the triple $(X_i,X_j,X_k)$ of variables interacts with
each of the three $\cdalg$ constraints $(\binmat ^P)_{ij}$, $(\binmat
^P)_{ik}$ and $(\binmat ^P)_{jk}$ on the pairs $(X_i,X_j)$,
$(X_i,X_k)$ and $(X_j,X_k)$. If $(\termat ^P)_{ijk}$ is an atomic
relation, say $r$, then the interaction is given by the three
functions $\fctroatocdaontw$, $\fctroatocdaonth$ and
$\fctroatocdatwth$ of Figure \ref{roa-comp-three}; namely:
\begin{enumerate}
  \item $(\binmat ^P)_{ij}\leftarrow\fctroatocdaontw (r,P,i,j,k)$;\\
    $(\binmat ^P)_{ji}\leftarrow ((\binmat ^P)_{ij})^\smile$;
  \item $(\binmat ^P)_{ik}\leftarrow\fctroatocdaonth (r,P,i,j,k)$;\\
    $(\binmat ^P)_{ki}\leftarrow ((\binmat ^P)_{ik})^\smile$;
  \item $(\binmat ^P)_{jk}\leftarrow\fctroatocdatwth (r,P,i,j,k)$;\\
    $(\binmat ^P)_{kj}\leftarrow ((\binmat ^P)_{jk})^\smile$;
\end{enumerate}
If $(\termat ^P)_{ijk}$ is a disjunctive, non atomic relation, say
$R$, then the interaction is the union of the interactions at the
atomic level; namely:
\begin{enumerate}
  \item $(\binmat ^P)_{ij}\leftarrow\displaystyle\bigcup _{r\in R}\fctroatocdaontw (r,P,i,j,k)$;\\
    $(\binmat ^P)_{ji}\leftarrow ((\binmat ^P)_{ij})^\smile$;
  \item $(\binmat ^P)_{ik}\leftarrow\displaystyle\bigcup _{r\in R}\fctroatocdaonth (r,P,i,j,k)$;\\
    $(\binmat ^P)_{ki}\leftarrow ((\binmat ^P)_{ik})^\smile$;
  \item $(\binmat ^P)_{jk}\leftarrow\displaystyle\bigcup _{r\in R}\fctroatocdatwth (r,P,i,j,k)$;\\
    $(\binmat ^P)_{kj}\leftarrow ((\binmat ^P)_{jk})^\smile$;
\end{enumerate}
\subsection{CSPs of cardinal direction relations on 2D points}
We define a $\cdalg$-CSP as a CSP of which the constraints are $\cdalg$
relations on pairs of the variables. The universe of a $\cdalg$-CSP is
the set $\BBR ^2$ of 2D points.

A $\cdalg$-matrix of order $n$ is a binary constraint matrix of order $n$ of
which the entries are $\cdalg$ relations. The constraint matrix
associated with a $\cdalg$-CSP is a $\cdalg$-matrix.

A scenario of a $\cdalg$-CSP is a refinement $P'$ such that all entries
of the constraint matrix of $P'$ are atomic relations.

If we make the assumption that a  $\cdalg$-CSP does not include the empty
constraint, which indicates a trivial inconsistency,
then a $\cdalg$-CSP is strongly $2$-consistent.
\subsection*{Solving a $\cdalg$-CSP}
A simple adaptation of Allen's constraint propagation algorithm \cite{Allen83b}
can be used to achieve path consistency (hence strong $3$-consistency)
for $\cdalg$-CSPs. Applied to a $\cdalg$-CSP $P$, such an adaptation would
repeat the following steps until either stability is reached or the empty
relation is detected (indicating inconsistency):
\begin{enumerate}
  \item Consider a triple $(X_i,X_j,X_k)$ of variables verifying
        $(\binmat ^P)_{ij}\not\subseteq (\binmat ^P)_{ik}\circ (\binmat ^P)_{kj}$
  \item $(\binmat ^P)_{ij}\leftarrow (\binmat ^P)_{ij}\cap (\binmat ^P)_{ik}\circ (\binmat ^P)_{kj}$
  \item If $((\binmat ^P)_{ij}=\emptyset)$ then exit (the CSP is inconsistent).
\end{enumerate}
Path consistency is complete for atomic $\cdalg$-CSPs
\cite{Ligozat98a}. Given this, Ladkin and Reinefeld's solution search
algorithm \cite{Ladkin92a} can be used to search for a solution, if
any, or otherwise report inconsistency, of a general $\cdalg$-CSP.
\subsection{CSPs on relative orientation of 2D points}
We define an $\roalg$-CSP as a CSP of which
the constraints are $\roalg$ relations on triples of the variables. The universe of an $\roalg$-CSP is
the set $\BBR ^2$ of 2D points.

An $\roalg$-matrix of order $n$ is a ternary constraint matrix of order $n$ of
which the entries are $\roalg$ relations. The constraint matrix
associated with an $\roalg$-CSP is an $\roalg$-matrix.

A scenario of an $\roalg$-CSP is a refinement $P'$ such that all entries
of the constraint matrix of $P'$ are atomic relations.

If we make the assumption that an  $\roalg$-CSP does not include the empty
constraint, which indicates a trivial inconsistency,
then an $\roalg$-CSP is strongly $3$-consistent.
\subsection*{Searching for a strongly  $4$-consistent scenario of an $\roalg$-CSP}
A simple adaptation of the constraint propagation algorithm in \cite{Isli00b}
can be used to achieve strong $4$-consistency for $\roalg$-CSPs. Applied to an
$\roalg$-CSP $P$, such an adaptation would repeat the following steps until either
stability is reached or the empty relation is detected (indicating inconsistency):
\begin{enumerate}
  \item Consider a quadruple $(X_i,X_j,X_k,X_l)$ of variables verifying
        $(\termat ^P)_{ijl}\not\subseteq (\termat ^P)_{ijk}\circ (\termat ^P)_{ikl}$
  \item $(\termat ^P)_{ijl}\leftarrow (\termat ^P)_{ijl}\cap (\termat ^P)_{ijk}\circ (\binmat ^P)_{ikl}$
  \item If $((\termat ^P)_{ijl}=\emptyset)$ then exit (the CSP is inconsistent).
\end{enumerate}
In \cite{Isli00b}, the authors have proposed a complete solution search algorithm
for CSPs expressed in their $\atra$ algebra. The algorithm is similar to the
one in \cite{Ladkin92a} for temporal interval networks \cite{Allen83b}, except
that:
\begin{enumerate}
  \item it refines the relation on a triple of variables at each node of the search
    tree, instead of the relation on a pair of variables; and
  \item it makes use of a constraint propagation procedure achieving strong
    $4$-consistency, in the preprocessing step and as the filtering method during
    the search, instead of a procedure achieving path consistency.
\end{enumerate}
Unless we can prove that the strong $4$-consistency
procedure in \cite{Isli00b} is complete for the $\roalg$ atomic relations, we cannot
claim completeness of the solution search procedure for general 
$\roalg$-CSPs. But we can still use the procedure to search for a
strongly $4$-consistent scenario of the input CSP.
For more details on the algorithm, and on its binary counterpart, the
reader is referred to \cite{Isli00b,Ladkin92a}.
\subsection{CSPs of cardinal direction relations and relative
  orientation relations on 2D points}
We define a $\combalg$-CSP as a CSP of which
the constraints consist of a conjunction of $\cdalg$ relations on
pairs of the variables, and $\roalg$ relations on triples of the
variables. The universe of a $\combalg$-CSP is the set $\BBR ^2$ of 2D points.
\subsection*{Matrix representation of a $\combalg$-CSP}
A $\combalg$-CSP $P$ can, in an obvious way, be represented as two
constraint matrices:
\begin{enumerate}
  \item a binary constraint matrix, $\binmat ^P$,
    representing the $\cdalg$ part of $P$, i.e., the subconjunction
    consisting of $\cdalg$ relations on pairs of the variables; and
  \item a ternary constraint matrix, $\termat ^P$, representing the $\roalg$
    part of $P$, i.e., the rest of the conjunction, consisting of
    $\roalg$ relations on triples of the variables.
\end{enumerate}
We refer to the representation as $\langle\binmat ^P,\termat ^P\rangle$.
The $\binmat ^P$ entry $(\binmat ^P )_{ij}$consists of the 
$\cdalg$ relation on the pair $(X_i,X_j)$ of variables. Similarly, the 
$\termat ^P$ entry $(\termat ^P)_{ijk}$ consists of the $\roalg$ relation on 
the triple $(X_i,X_j,X_k)$ of variables.
\subsection*{A constraint propagation procedure for $\combalg$-CSPs}
A path consistency algorithm, such as the one in \cite{Allen83b}, applied 
to a binary CSP such as a $\cdalg$-CSP, uses a queue $\queue$, which
can be supposed, for simplicity, to have been initialised to all pairs 
$(x,y)$ of the CSP variables verifying $x\leq y$ (the variables are
supposed to be ordered). The algorithm removes one pair of variables from
$\queue$ at a time; a removed pair is used to eventually update the
relations on the neighbouring pairs of variables (pairs sharing at
least one variable). Whenever such a pair is successfully updated, it
is entered into $\queue$, if it is not already there, in order to be
considered at a future stage for propagation. The algorithm
terminates if the empty relation, indicating inconsistency, is
detected, or if $\queue$ becomes empty, indicating that a fixed 
point has been reached and the input CSP is made path consistent.

A strong $4$-consistency algorithm, such as the one in
\cite{Isli00b}, applied to a ternary CSP such as an
$\roalg$-CSP, is, somehow, an adaptation to ternary relations of a
path consistency algorithm. It uses a queue $\queue$, which can be
supposed, for simplicity, to have been initialised to all triples
$(x,y,z)$ of the CSP variables such that $x\leq y\leq z$. The
algorithm removes one triple from $\queue$ at a time; a removed
triple is used to eventually update the relations on the neighbouring
triples (sharing at least two variables). Whenever such a triple is
successfully updated, it is entered into $\queue$, if it is not
already there, in order to be considered at a future stage for
propagation. The algorithm
terminates if the empty relation, indicating inconsistency, is
detected, or if $\queue$ becomes empty, indicating that a fixed 
point has been reached and the input CSP is made strongly $4$-consistent.

In Figure \ref{strong4consistency}, we propose a constraint propagation procedure,
$\pcsfc$, for $\combalg$-CSPs, which aims at:
\begin{enumerate}
  \item achieving \underline{p}ath \underline{c}onsistency ($\mbox{\em
      Pc}$) for the $\cdalg$ projection, using, for instance, the
    algorithm in \cite{Allen83b};
  \item achieving \underline{s}trong
    $\underline{4}$-\underline{c}onsistency
    ($\mbox{\em S4c}$) for the $\roalg$ projection, using, for
    instance, the algorithm in \cite{Isli00b}; and
  \item  more ($\mbox{\em +}$).
\end{enumerate}
The procedure does more than just achieving path consistency for the
$\cdalg$ projection, and strong $4$-consistency for the $\roalg$
projection. It implements as well the interaction between the two
combined calculi; namely:
\begin{enumerate}
  \item The path consistency operation,
    $(\binmat ^P)_{ik}\leftarrow (\binmat ^P)_{ik}\cap (\binmat
    ^P)_{ij}\circ (\binmat ^P)_{jk}$,
    which, under normal circumstances, operates internally, within a
    same CSP, should now be, and is, augmented so that it can send
    information from the $\cdalg$
    component into the $\roalg$ component; this is achieved by a call
    to the procedure $\pairprop ()$. Specifically, whenever a pair
    $(X_i,X_j)$ of variables is taken from $\queue$ for propagation,
    the following is performed for all variables $X_k$:
   \begin{enumerate}
     \item[$\bullet$] the procedure $\pairprop ()$ of Figure
       \ref{strong4consistency} checks whether the relation on the
       pair $(X_i,X_k)$ ---see lines \ref{pp1}-\ref{pp4}--- or the
       relation on the pair $(X_k,X_j)$ ---see lines
       \ref{pp5}-\ref{pp8}--- can be successfully updated. If this
       happens, the corresponding pairs of variables are
       entered into $\queue$ in order to be considered for
       propagation at a later point of the process. This part
       of the propagation is not new, and is widely known in the
       literature on propagation algorithms, such as path
       consistency (see \cite{Allen83b} for the case of constraint-based
       qualitative temporal reasoning). What is new in the procedure
       $\pairprop ()$ is the call to the procedure $\cdatoroa ()$
       ---see lines \ref{pp4p} and \ref{pp9}--- which aims at checking, whenever a pair
       $(X_i,X_j)$ is taken from $\queue$, whether the $\cdalg$
       relation on $(X_i,X_j)$ can
       update the $\roalg$ relation on the triple $(X_i,X_j,X_k)$ or that on the triple $(X_k,X_i,X_j)$. If
       either of the two $\roalg$ relations gets successfully updated, the corresponding triple
       of variables is entered into $\queue$ in order to be considered
       for propagation at a later point of the process. The
       procedure $\cdatoroa ()$ is the implementation of the $\cdatoroa$
       interaction operation, $\otimes$, defined in the table of Figure \ref{cda-comp}, which
       outputs the $\roalg$ relation, $r\inferred s$, logically
       implied by the conjunction of two $\cdalg$ atoms,
       $r$ and $s$.
   \end{enumerate}
  \item The strong $4$-consistency operation,
    $(\termat ^P)_{ijk}\leftarrow (\termat ^P)_{ijk}\cap (\termat
    ^P)_{ijl}\circ (\termat ^P)_{ilk}$,
    which also operates internally under normal circumstances, is
    augmented so that it can send information from the $\roalg$
    component into the $\cdalg$ component; this is achieved by a call
    to the procedure $\tripleprop ()$. Specifically, whenever a triple
    $(X_i,X_j,X_k)$ is taken from $\queue$ for propagation, the
    following is performed for all variables $X_m$:
   \begin{enumerate}
     \item[$\bullet$] the procedure $\tripleprop ()$ of Figure
       \ref{strong4consistency} checks whether the relation on the
       triple $(X_i,X_j,X_m)$ ---see lines \ref{tp1}-\ref{tp4}--- or the
       relation on the triple $(X_i,X_k,X_m)$ ---see lines
       \ref{tp5}-\ref{tp8}--- or the
       relation on the triple $(X_j,X_k,X_m)$ ---see lines
       \ref{tp9}-\ref{tp12}--- can be successfully updated. If this
       happens, the corresponding triples of variables are
       entered into $\queue$ in order to be considered for
       propagation at a later point of the process. This part
       of the propagation is taken from the strong $4$-consistency
       algorithm in \cite{Isli00b}. What is new in the procedure
       $\tripleprop ()$ is the call to the procedure $\roatocda ()$
       ---see line \ref{tp13}--- which aims at checking, whenever a triple
       $(X_i,X_j,X_k)$ is taken from $\queue$, whether the $\roalg$
       relation on $(X_i,X_j,X_k)$ can
       update the $\cdalg$ relations on the different pairs of the three
       arguments: the pairs $(X_i,X_j)$, $(X_i,X_k)$ and $(X_j,X_k)$. If
       any of the three $\cdalg$ relations gets successfully updated, the
       corresponding pair of variables is entered into $\queue$ in order to be considered
       for propagation at a later point of the process. The
       procedure $\roatocda ()$ is the implementation of the $\roatocda$ interaction table of Figure \ref{roa-comp-three}.
   \end{enumerate}
\end{enumerate}
\begin{figure}
\begin{scriptsize}
\begin{enumerate}
  \item[] Input: the matrix representation $\langle\binmat ^P,\termat ^P\rangle$
    of a $\combalg$-CSP $P$ with set of variables $V$.
  \item[] Output: the CSP $P$ made strongly $4$-consistent.
  \item[]      {\em procedure} $\pcsfc$;
  \item      initialise $\queue$: $\queue\leftarrow\{(x,y)\in V^2:x\leq
    y\}\cup\{(x,y,z)\in V^3:x\leq y\leq z\}$;
  \item      repeat\{
  \item      \hskip 0.2cm get (and remove) next element $Q$ from
    $Queue$;
  \item      \hskip 0.2cm if $Q$ is a pair, say $(X_i,X_j)$\{
  \item      \hskip 0.4cm for $k\leftarrow 1$ to $n$\{$\pairprop (P,i,j,k)$;\}
  \item      \hskip 0.4cm \}
  \item      \hskip 0.2cm else ($Q$ is a triple, say $(X_i,X_j,X_k)$)\{
  \item      \hskip 0.4cm for $m\leftarrow 1$ to $n$\{$\tripleprop (P,i,j,k,m)$;\}
  \item      \hskip 0.4cm \}
  \item     \hskip 0.2cm \}
  \item     until $Queue$ is empty;
\end{enumerate}
\begin{enumerate}
  \item[]      {\em procedure} $\pairprop (P,i,j,k)$;
  \item\label{pp1}      \hskip 0.8cm $\temp\leftarrow (\binmat ^P)_{ik}\cap (\binmat ^P)_{ij}\circ (\binmat ^P)_{jk}$;
  \item      \hskip 0.8cm If $\temp =\emptyset$ then exit (the CSP is
            inconsistent);
  \item      \hskip 0.8cm if $\temp\not = (\binmat ^P)_{ik}$
  \item\label{pp4}      \hskip 1.2cm \{add-to-queue$(X_i,X_k)$;
                            $(\binmat ^P)_{ik}\leftarrow\temp ;$
                            $(\binmat ^P)_{ki}\leftarrow\temp ^\smile
                            ;$
                          \}
  \item\label{pp4p}      \hskip 0.8cm $\cdatoroa (P,i,j,k)$;
  \item\label{pp5}      \hskip 0.8cm $\temp\leftarrow (\binmat ^P)_{kj}\cap (\binmat ^P)_{ki}\circ (\binmat ^P)_{ij}$;
  \item      \hskip 0.8cm If $\temp =\emptyset$ then exit (the CSP is
            inconsistent);
  \item      \hskip 0.8cm if $\temp\not = (\binmat ^P)_{kj}$
  \item\label{pp8}      \hskip 1.2cm \{add-to-queue$(X_k,X_j)$;
                            $(\binmat ^P)_{kj}\leftarrow\temp ;$
                            $(\binmat ^P)_{jk}\leftarrow\temp ^\smile ;$
                          \}
  \item\label{pp9}      \hskip 0.8cm $\cdatoroa (P,k,i,j)$;
\end{enumerate}
\begin{enumerate}
  \item[]      {\em procedure} $\tripleprop (P,i,j,k,m)$;
  \item\label{tp1}      \hskip 0.8cm $Temp\leftarrow (\termat ^P)_{ijm}\cap (\termat ^P)_{ijk}\circ (\termat ^P)_{ikm}$;
  \item      \hskip 0.8cm If $Temp=\emptyset$ then exit (the CSP is
            inconsistent);
  \item      \hskip 0.8cm if $Temp\not = (\termat ^P)_{ijm}$
  \item\label{tp4}      \hskip 1.2cm \{add-to-queue$(X_i,X_j,X_m)$;$update(P,i,j,m,Temp)$;\}
  \item\label{tp5}      \hskip 0.8cm $Temp\leftarrow (\termat ^P)_{ikm}\cap (\termat ^P)_{ikj}\circ (\termat ^P)_{ijm}$;
  \item     \hskip 0.8cm If $Temp=\emptyset$ then exit (the CSP is
            inconsistent);
  \item     \hskip 0.8cm if $Temp\not = (\termat ^P)_{ikm}$
  \item\label{tp8}     \hskip 1.2cm \{add-to-queue$(X_i,X_k,X_m)$;$update(P,i,k,m,Temp)$;\}
  \item\label{tp9}     \hskip 0.8cm $Temp\leftarrow (\termat ^P)_{jkm}\cap (\termat ^P)_{jki}\circ (\termat ^P)_{jim}$;
  \item     \hskip 0.8cm If $Temp=\emptyset$ then exit (the CSP is
            inconsistent);
  \item     \hskip 0.8cm if $Temp\not = (\termat ^P)_{jkm}$
  \item\label{tp12}     \hskip 1.2cm \{add-to-queue$(X_j,X_k,X_m)$;$update(P,j,k,m,Temp)$;\}
  \item\label{tp13}     \hskip 0.8cm $\roatocda (P,i,j,k)$;
\end{enumerate}
\begin{enumerate}
  \item[]      {\em procedure} $update(P,i,j,k,T)$;
  \item $(\termat ^P)_{ijk}\leftarrow T;
         (\termat ^P)_{ikj}\leftarrow T^\smile ;
         (\termat ^P)_{jki}\leftarrow T ^\frown ;$
  \item $(\termat ^P)_{jik}\leftarrow ((\termat ^P)_{jki})^\smile ;
         (\termat ^P)_{kij}\leftarrow ((\termat ^P)_{jki})^\frown ;
         (\termat ^P)_{kji}\leftarrow ((\termat ^P)_{kij})^\smile ;$
\end{enumerate}
\caption{A constraint propagation procedure, $\pcsfc$, for
  $\combalg$-CSPs. The procedures $\cdatoroa$ and $\roatocda$ used by
  the algorithm are defined in Figure \ref{interaction}.}\label{strong4consistency}
\end{scriptsize}
\end{figure}
\begin{figure}
\begin{scriptsize}
\begin{enumerate}
  \item[]      {\em procedure} $\cdatoroa (P,i,j,k)$;
  \item      \hskip 0.8cm $\roair\leftarrow\displaystyle\bigcup
    _{r_1\in (\binmat ^P)_{ij},r_2\in (\binmat ^P)_{jk}}r_1\otimes r_2$;
  \item      \hskip 0.8cm $Temp\leftarrow (\termat ^P)_{ijk}\cap\roair$;
  \item      \hskip 0.8cm If $Temp=\emptyset$ then exit (the CSP is
            inconsistent);
  \item      \hskip 0.8cm if $Temp\not = (\termat ^P)_{ijk}$
  \item      \hskip 1.2cm \{add-to-queue$(X_i,X_j,X_k)$;$update(P,i,j,k,Temp)$;\}
\end{enumerate}
\begin{enumerate}
  \item[]      {\em procedure} $\roatocda (P,i,j,k)$;
  \item      \hskip 0.8cm $Temp\leftarrow\displaystyle\bigcup _{r\in R}\fctroatocdaontw (r,P,i,j,k)$;
  \item      \hskip 0.8cm If $Temp=\emptyset$ then exit (the CSP is inconsistent);
  \item      \hskip 0.8cm if $Temp\not = (\termat ^P)_{ij}$
  \item      \hskip 1.2cm \{add-to-queue$(X_i,X_j)$;
                            $(\binmat ^P)_{ij}\leftarrow\temp ;$
                            $(\binmat ^P)_{ji}\leftarrow\temp ^\smile ;$
                          \}
  \item      \hskip 0.8cm $Temp\leftarrow\displaystyle\bigcup _{r\in R}\fctroatocdaonth (r,P,i,j,k)$;
  \item      \hskip 0.8cm If $Temp=\emptyset$ then exit (the CSP is inconsistent);
  \item      \hskip 0.8cm if $Temp\not = (\termat ^P)_{ik}$
  \item      \hskip 1.2cm \{add-to-queue$(X_i,X_k)$;
                            $(\binmat ^P)_{ik}\leftarrow\temp ;$
                            $(\binmat ^P)_{ki}\leftarrow\temp ^\smile ;$
                          \}
  \item      \hskip 0.8cm $Temp\leftarrow\displaystyle\bigcup _{r\in R}\fctroatocdatwth (r,P,i,j,k)$;
  \item      \hskip 0.8cm If $Temp=\emptyset$ then exit (the CSP is inconsistent);
  \item      \hskip 0.8cm if $Temp\not = (\termat ^P)_{jk}$
  \item      \hskip 1.2cm \{add-to-queue$(X_j,X_k)$;
                            $(\binmat ^P)_{jk}\leftarrow\temp ;$
                            $(\binmat ^P)_{kj}\leftarrow\temp ^\smile ;$
                          \}
\end{enumerate}
\caption{The procedures $\cdatoroa$ and $\roatocda$ used by the
  constraint propagation algorithm $\pcsfc$ of Figure \ref{strong4consistency}.}\label{interaction}
\end{scriptsize}
\end{figure}
\begin{thr}\label{theorem1}
The constraint propagation procedure $\pcsfc$ runs into completion in $O(n^4)$ time, where
$n$ is the number of variables of the input $\combalg$-CSP.
\end{thr}
{\bf Proof. }
The number of variable pairs is  $O(n^2)$, whereas the number of variable
triples is $O(n^3)$. A pair as well as a triple may be placed in
Queue at most a constant number of times (9 for a pair, which is the total number of
$\cdalg$ atoms; and also 9 for a triple, which is the total number of
$\roalg$ atoms). Every time a pair or a triple is removed from Queue for
propagation, the procedure performs $O(n)$ operations. \cqfd
\begin{ex}
Consider again the description of Example \ref{example1}. We can
represent the situation as a $\combalg$-CSP with variables $X_b$,
$X_h$, $X_l$ and $X_p$, standing for the cities of Berlin, Hamburg,
London and Paris, respectively.
\begin{enumerate}
  \item The knowledge "viewed from Hamburg, Berlin is to the left of
    Paris" translates into the $\roalg$ constraint $\leftn
    (X_h,X_p,X_b)$: $(\termat ^P)_{hpb}=\{\leftn\}$.
  \item The other $\roalg$ knowledge translates as follows:
        $(\termat ^P)_{hlp}=\{\leftn\}$,
        $(\termat ^P)_{hlb}=\{\leftn\}$,
        $(\termat ^P)_{lpb}=\{\leftn\}$.
  \item The $\cdalg$ part of the knowledge translates as follows:
        $(\binmat ^P)_{hp}=\{\north\}$,
        $(\binmat ^P)_{hb}=\{\northwest\}$,
        $(\binmat ^P)_{pl}=\{\south\}$.
\end{enumerate}
As discussed in Example \ref{example1}, reasoning separately about the 
two components of the knowledge shows two consistent components,
whereas the combined knowledge is clearly inconsistent. Using the
procedure $\pcsfc$, we can detect the inconsistency in the following
way. From the $\cdalg$ constraints $(\binmat ^P)_{hp}=\{\north\}$ and
        $(\binmat ^P)_{pl}=\{\south\}$, the algorithm infers, using
        the augmented $\cdalg$ composition table of Figure
        \ref{cda-comp} ---specificaly, the $\cdatoroa$ interaction
        operation $\otimes$--- the 
        $\roalg$ relation $\{\behonen ,\coinconen ,\betwn\}$ on the
        triple $(X_h,X_p,X_l)$. The conjunction of the inferred knowledge
        $\{\behonen ,\coinconen ,\betwn\}(X_h,X_p,X_l)$ and the already
        existing knowledge $\{\leftn\}(X_h,X_l,X_p)$ ---equivalent to
        $\{\rightn\}(X_h,X_p,X_l)$--- gives the empty relation,
        indicating the inconsistency of the knowledge.
\end{ex}
\section{Discussion}\label{discussion}
Current research shows clearly the importance of developing
spatial RAs: specialising an $\alcd$-like Description Logic (DL) \cite{Baader91a}, so that the
roles are temporal immediate-successor (accessibility) relations, and the concrete domain is generated by a
decidable spatial RA in the style of the well-known Region-Connection Calculus RCC-8 \cite{Randell92a},
leads to a computationally well-behaving family of
languages for spatial change in general, and for motion of spatial scenes in particular:
\begin{enumerate}
  \item Deciding satisfiability of an $\alcd$ concept $\wrt$ to a cyclic TBox is, in
    general, undecidable (see, for instance, \cite{Lutz04a}).
  \item In the case of the spatio-temporalisation, however, if we use what is called
    weakly cyclic TBoxes in \cite{Isli02d}, then
    satisfiability of a concept $\wrt$ such a TBox is decidable. The axioms of a
    weakly cyclic TBox capture the properties of modal temporal operators. The
    reader is referred to \cite{Isli02d} for details.
\end{enumerate}
Spatio-temporal theories such as the ones defined in \cite{Isli02d}
can be seen as single-ontology spatio-temporal theories, in the sense
that the concrete domain represents only one type of spatial knowledge 
(e.g., \mbox{RCC-8} relations if the concrete domain is generated by
\mbox{RCC-8}). We could extend such theories to handle more than just
one concrete domain: for instance, two concrete domains, one generated
by $\cdalg$, the other by $\roalg$. This would lead to what could be
called multi-ontolopgy spatio-temporal theories. The presented work
clearly shows that the reasoning issue in such multi-ontology
theories does not reduce to reasoning about the projections onto the
different concrete domains.

Before we provide an example, we adapt a definition from \cite{Isli02d}. $\xdlzo$
is obtained from $\alcd$ by temporalisng the roles, and spatialising the concrete domain: $\xdlzo$ has exactly one role which is functional, and which we refer to in the following as $f$ (the subscript $0$ indicates the number of general, not necessarily functional roles, and the subscript $1$ the number of functional roles).
The roles in $\alc$, as well as the roles other than the abstract features in $\alcd$, are
interpreted in a similar way as the modal operators of the multi-modal logic
${\cal K}_{(m)}$ \cite{Halpern85a}. A functional role is also referred to as an abstract feature. $f$ plays the role of the NEXT operator in linear time temporal logic: $f$ is antisymmetric, serial and linear.
\begin{definition}[$\xdlzo$ concepts]\label{defxdlconcepts}
Let $N_C$ and $N_{cF}$ be mutually disjoint and countably infinite sets of concept
names and concrete features, respectively. A (concrete)
feature chain is any finite composition $f_1\ldots f_ng$ of $n\geq 0$ abstract
features $f_1,\ldots ,f_n$ and one concrete feature $g$. The set of $\xdlzo$
concepts is the smallest set such that:
\begin{enumerate}
  \item\label{defxdlconceptsone} $\top$ and $\bot$ are $\xdlzo$ concepts
  \item\label{defxdlconceptstwo} an $\xdlzo$ concept name is an $\xdlzo$
    (atomic) concept
  \item\label{defxdlconceptsthree} if
    $C$ and $D$ are $\xdlzo$ concepts;
    $g$ is a concrete feature;
    $u_1$ and $u_2$ are feature chains; and
    $P$ is an $\xdlzo$ predicate,\footnote{A predicate is any $\combalg$ relation.}
    then the following expressions are also $\xdlzo$ concepts:
    \begin{enumerate}
      \item\label{defxdlconceptsthreea} $\neg C$,
            $C\sqcap D$,
            $C\sqcup D$,
            $\exists f.C$,
            $\forall f.C$; and
      \item\label{defxdlconceptsthreeb}
            $\exists (u_1)(u_2).P$.
    \end{enumerate}
\end{enumerate}
\end{definition}
\begin{example}[illustration of $\xdlzo$]\label{xdlzocda}
\begin{figure*}[t]
\epsffile{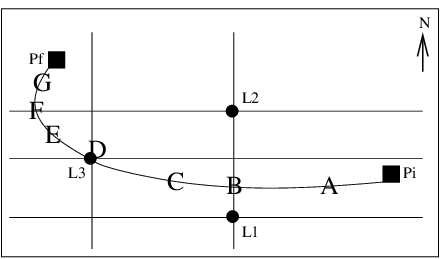}
\caption{Illustration of $\xdlzo$: the
upward arrow pointing at N indicates North.}\label{sectorstwo}
\end{figure*}
Consider a satellite-like high-level surveillance system, aimed at the 
surveillance of flying aeroplanes within a three-landmark environment.
The basic task of the system is to situate qualitatively an aeroplane
relative to the different landmarks, as well as to relate
qualitatively the different positions of an aeroplane while in flight.
If the system is used for the surveillance of the European sky, the
landmarks could be capitals of European countries, such as Berlin,
London and Paris. For the purpose, the system uses a high-level
spatial description language, such as a QSR language, which we
suppose in this example to be the $\combalg$ calculus defined in this work.
The example is illustrated in Figure \ref{sectorstwo}. The horizontal and
vertical lines through the three
landmarks partition the plane into 0-, 1- and 2-dimensional regions, as shown
in Figure \ref{sectorstwo}. The flight of an aeroplane within the
environment, 
starts from some point $P_i$ in Region $A$ (initial region), and ends at some
point $P_f$ in Region $G$ (final, or goal region). Immediately after the
initial region, the flight ``moves'' to Region $B$, then to Region $C$,
$\ldots$, then to Region $F$, and finally to the goal region $G$. The
surveillance system has the task of providing qualitative knowledge on how it
``sees'' the aeroplane at each moment of the flight, knowledge consisting of
$\combalg$ (i.e., $\cdalg$ and $\roalg$) relations. The whole knowledge consists mainly of a recording
of successive snapshots of the flight, one per region.
The $\cdalg$ component of
a snapshot is a conjunction of constraints giving, for instance, the $\cdalg$ relation
relating the aeroplane to each of the three landmarks. 
The $\roalg$ component of the knowledge provides, for instance, the $\roalg$ relation on triples of the aeroplane's positions at the different regions. The entire flight consists of a succession of
subflights, $f_A,f_B,\ldots ,f_G$,
such that $f_B$ immediately follows $f_A$, $f_C$ immediately follows $f_B$,
$\ldots$, and $f_G$ immediately follows $f_F$. Subflight $f_X$,
$X\in\{A,\ldots ,G\}$, takes place in Region $X$, and gives rise to a
defined concept $B_X$ describing the panorama of the aeroplane O while in Region $X$, and
saying which subflight takes place next, i.e., which Region is
flied over next. We make use of the concrete features $g_{l1}$, $g_{l2}$,
$g_{l3}$ and $g_{o}$, which have the task of ``referring'', respectively,  to
the actual positions of landmarks $l_1$, $l_2$, $l_3$, and of the aeroplane
$O$. As roles, the unique (functional) role of $\xdlzo$, referred to as $f$,
and denoting the linear-time immediate successor function. The acyclic TBox
composed of the following axioms describes the flight:
\begin{footnotesize}
\begin{eqnarray}
B_A&\doteq&
           \exists (g_{o})(g_{l1}).\northeast\sqcap
           \exists (g_{o})(g_{l2}).\southeast\sqcap
           \exists (g_{o})(g_{l3}).\southeast\sqcap
           \exists f.B_B  \nonumber\\
B_B&\doteq&
           \exists (g_{o})(g_{l1}).\north\sqcap
           \exists (g_{o})(g_{l2}).\south\sqcap
           \exists (g_{o})(g_{l3}).\southeast\sqcap
           \exists f.B_C  \nonumber\\
B_C&\doteq&
           \exists (g_{o})(g_{l1}).\northwest\sqcap
           \exists (g_{o})(g_{l2}).\southwest\sqcap
           \exists (g_{o})(g_{l3}).\southeast\sqcap
           \exists f.B_D  \nonumber\\
B_D&\doteq&
           \exists (g_{o})(g_{l1}).\northwest\sqcap
           \exists (g_{o})(g_{l2}).\southwest\sqcap
           \exists (g_{o})(g_{l3}).\equal\sqcap
           \exists f.B_E  \nonumber\\
B_E&\doteq&
           \exists (g_{o})(g_{l1}).\northwest\sqcap
           \exists (g_{o})(g_{l2}).\southwest\sqcap
           \exists (g_{o})(g_{l3}).\northwest\sqcap
           \exists f.B_F  \nonumber\\
B_F&\doteq&
           \exists (g_{o})(g_{l1}).\northwest\sqcap
           \exists (g_{o})(g_{l2}).\west\sqcap
           \exists (g_{o})(g_{l3}).\northwest\sqcap
           \exists f.B_G  \nonumber\\
B_G&\doteq&
           \exists (g_{o})(g_{l1}).\northwest\sqcap
           \exists (g_{o})(g_{l2}).\northwest\sqcap
           \exists (g_{o})(g_{l3}).\northwest  \nonumber
\end{eqnarray}
\end{footnotesize}
The concept $B_A$, for instance, describes the snapshot of the plane while in Region $A$.
It says that the aeroplane is
                           northeast landmark $L_1$ ($\exists (g_{o})(g_{l1}).\northeast$);
                           southeast landmark $L_2$ ($\exists (g_{o})(g_{l2}).\southeast$); and
                           southeast landmark $L_3$ ($\exists (g_{o})(g_{l3}).\southeast$).
The concept also says that the subflight to take place next is $f_B$ ($\exists f.B_B$).

One might want as well the system to provide $\cdalg$ knowledge on how the aeroplane's different positions during the
flight relate to each other. For example, that the aeroplane, while in region C, remains northwest of its position
while in region B; or, that the position, while in the goal region G, remains northwest of the position
while in region E. These two constraints can be injected into the TBox by modifying
the axioms $B_B$ and $B_E$ as follows:
\begin{footnotesize}
\begin{eqnarray}
B_B&\doteq&
           \exists (g_{o})(g_{l1}).\north\sqcap
           \exists (g_{o})(g_{l2}).\south\sqcap
           \exists (g_{o})(g_{l3}).\southeast\sqcap
           \exists (g_{o})(fg_{o}).\southeast\sqcap
           \exists f.B_C  \nonumber\\
B_E&\doteq&
           \exists (g_{o})(g_{l1}).\northwest\sqcap
           \exists (g_{o})(g_{l2}).\southwest\sqcap
           \exists (g_{o})(g_{l3}).\northwest\sqcap
           \exists (g_{o})(ffg_{o}).\southeast\sqcap
           \exists f.B_F  \nonumber
\end{eqnarray}
\end{footnotesize}
So far, the example has made use of $\cdalg$ relations only as predicates. One
might want to represent knowledge such as, the flight from Region B until
Region D had a clockwise curvature (i.e., the aeroplane, while in region C,
kept to the left of the directed line joining the position while in Region B
to the position while in Region D). Another kind of knowledge one might want to
represent is that, the flight was rectilinear in Region E. These can be added
to the existing knowledge by modifying defined concepts $B_B$ and $B_D$ as
follows:
\begin{footnotesize}
\begin{eqnarray}
B_B&\doteq&
           \exists (g_{o})(g_{l1}).\north\sqcap
           \exists (g_{o})(g_{l2}).\south\sqcap
           \exists (g_{o})(g_{l3}).\southeast\sqcap
           \exists (g_{o})(fg_{o}).\southeast\sqcap
           \exists (g_{o})(ffg_{o})(fg_{o}).\leftn\sqcap
           \exists f.B_C  \nonumber\\
B_D&\doteq&
           \exists (g_{o})(g_{l1}).\northwest\sqcap
           \exists (g_{o})(g_{l2}).\southwest\sqcap
           \exists (g_{o})(g_{l3}).\equal\sqcap
           \exists (g_{o})(ffg_{o})(fg_{o}).\betwn\sqcap
           \exists f.B_E  \nonumber
\end{eqnarray}
\end{footnotesize}
\end{example}
\section{Summary}\label{summary}
We have presented the integration of two calculi of spatial relations
well-known in Qualitative Spatial Reasoning (QSR): the
projection-based cardinal direction calculus in \cite{Frank92b},
and a coarser version of the relative orientation calculus in
\cite{Freksa92b}. With a $\gis$ example, we have shown
that reducing the issue of reasoning about knowledge expressed in the
integrating language to a simple matter of reasoning separately about each
of the two components was not sufficient. In other words, the interaction between the
two kinds of knowledge has to be handled: we have provided a
constraint propagation algorithm for such a purpose, which:
\begin{enumerate}
  \item achieves path consistency for the cardinal direction component;
  \item achieves strong $4$-consistency for the relative orientation
    component; and
  \item implements the interaction between the two kinds of knowledge.
\end{enumerate}

Integrating different kinds of knowledge is an emerging
and challenging issue in QSR. Similar work could be carried out for other aspects of knowledge in
QSR, such as qualitative distance \cite{Clementini97b} and relative
orientation \cite{Freksa92b}, an integration known to be
highly important for $\gis$ and robot navigation applications, and on
which not much has been achieved so far.
\begin{tiny}
\bibliographystyle{plain}
\bibliography{}
\end{tiny}
\newpage\noindent
\begin{center}
THE AMAI NOTIFICATION LETTER\\
(as received on 31 August 2004)
\end{center}
Dear Amar,

We are sorry to inform you that the paper you submitted to the
special volume of the Annals of Mathematics and Artificial
Intelligence dedicated to the 2004 AI+Math Symposium,
"Integrating cardinal direction relations in QSR",
has not been accepted for publication.

We have attached the reviews of the paper below.

We thank you for your efforts in preparing this submission,
and hope that you will find an appropriate forum for this work.

Sincerely,

Special Issue Co-Editors

=================REVIEWER 1======================

The work presented in this paper concerns the qualitative
spatial reasoning (QSR). A formalism (called cCOA) combining the
cardinal direction calculus (CDA) and the relative orientation
calculus (ROA) is studied. In this formalism the spatial
information about relative positions between the objects is
represented by two CSPs: a CSP whose constraints are defined
with relations of CDA (a CDA CSP) and a second CSP defined
with relations of ROA (a ROA CSP). Such a pair is called a
cCOA CSP. The main contribution of the paper consists in the
definition of a constraint propagation  algorithm allowing to
test the consistency of a cCOA CSP. A section of the paper is
devoted to the using of description logics with a concrete
domain (ALC(D)) to handle spatio-temporal information with
concrete domains generated from CDA and ROA.

\underline{Comments and recommendation}

Main comments :

1. Personally, I think that the paper is not very well structured. In particular,
I would displace the background on CSP after the presentation of the calculi and
before the definition of the cCOA CSP. Moreover, the cCOA CSP  are used in
Section 5.2 and Section 5.3 but defined in Section 5.5.

2. I found that numerous parts and sections could be enriched with relevant and
necessary details. For example, consider Section 2.3 which concerns Relation
Algebras. The content of this section is empty and does  bring anything to the
reader. As another example, consider Section 3 where the CDA calculus and the
ROA calculus are presented. In this section, we have just the names of the
relations of these calculus and the intuitive definitions of them. The minimum
thing would be to formally define these relations. There is other examples
through the paper.

3. To me, the fundamental problem of this paper is its contribution. Actually,
the main contribution of the paper consists in the definition of a constraint
propagation  procedure allowing to test the consistency of a cCOA CSP. This
algorithm is based on the usual path-consistency method which is improved by
operations making the translation of CDA relations to ROA relations and the
translation of ROA relations to CDA relations. This type of interaction between
two kinds of relations is not new, it has been already used for CSP containing
both qualitative and quantitative constraints. To be published in a journal the
paper must be enriched by a fundamental and relevant result such as the proof
that the proposed algorithm is complete when the constraints are atomic
relations or own another property.

4. Section 6 should be removed, I don't see the direct connection between this
section and the other sections.

Other comments :

Page 6, Section 2.2 : the 3-consistency and the path-consistency are the same
thing in the case where the constraints are binary constraints.

Page 7, line +1, the author must give some explanations about  the different
symbols, what is I for example ? Actually, the section 2.3 could be deleted.

Page 9, Section 4, What is the motivation to define a coarser relative
orientation calculus ? In the sequel, what is the calculus used : the initial
ROA calculus or the coarser ROA calculus ?

The author must give the order of the arguments for the relations lr, ll,
bw, ... For example, take the case where  A is between B and C, what is the
notation: bw(A,B,C) or bw(A,C,B) or ...

Page 11, the given definition of the composition  is not clear for me.
It seems that Lir(r)=Rir(r~). Is it exact ?

The functions roa-to-cdaxx use the CDA and ROA constraints to compute some new
CDA constraints. Consequently, the interaction is from CDA+ROA to CDA and not
from ROA to CDA.

Page 15, just before Section 5.5. : you must give the details (the proofs for
example) of these claims.

Page 19, Section 6 : the title of this section is not appropriated with its
content. Moreover, the topic of Section 6 is not directly connected with the
rest of the paper. I would remove this section.

Recommendation :

I cannot recommend to accept the paper in this form. I suggest
to the author to improve the paper in taking into account the
previous comments  before submitting again.

=================REVIEWER 2======================

Title: Integrating cardinal direction relations and other orientation relations
in Qualitative Spatial Reasoning

Author: Amar Isli

Overall evaluation: reject with encouragement to resubmit

\underline{GENERAL COMMENTS}

This paper presents a very interesting (and potentially useful) model of
qualitative spatial reasoning. This model expresses directional relations on
points by combining the binary relations of Frank and the ternary relations of
Freksa. Then, it presents a constraint propagation algorithm for the combined
model.

My major comments are as follows:

- Related work.

The author did not present related work at all. They only presented briefly and
very unclearly the adopted models. This must be fixed. The author should
present and discuss the models of Goyal and Egenhofer [Goyal00] and Billen and
Clementini [Billen04]. These model defined binary and ternary relations on
points and extended regions. Inference algorithms for the above models appear
in [Skiadopoulos04], [Skiadopoulos02] and [Billen04b]. References to the above
work should be also included.

- Presentation

The language of this paper is satisfactory (see also detailed comments). But
the presentation of this paper must be improved. The presentation of the model
is not complete. There are a lot of questions that are not answered. I believe
that the author should present formal definitions for this model. Moreover,
some examples would definitely help the reader.

- Contribution

The model is interesting but by itself it does not constitute a major
contribution. It is a simple combination of two existing models. On the other
hand, inference procedures are generally interesting and useful. In the present
paper the inference procedure is based on a transitivity table. There is no
discussion on whether this mechanism is complete or not. I can see that it
finds inconsistences that the inference procedures in the binary and ternary
model (alone) can not but does it work in EVERY case?

To conclude, I believe that Section 3 should be a section discussing related
work and maybe moved earlier. Then Section 4, should be extended, so that it
would describe and exemplify in DETAIL the proposed model. Finally the authors
should discuss (or better prove) whether their method is complete or not.

\underline{DETAILED COMMENTS}

Abstract, Line 6, "to the well-known poverty conjecture": Which conjecture?
Please explain and give references.

Page 2, Line 5, "the reader is referred to [4]": This is a 1997 survey. Since
then, there have been proposed several interesting models (both for points and
extended regions) for directional information (see discussion above). The
author should discuss these models and give the appropriate references.

Page 2, Paragraph 2, "An example on ...": I could not spot this example in this
paper. Moreover, I could not see the interaction and the integration of the two
calculi.

Page 2, Paragraph 3: This is an one-paragraph sentence. It is very difficult to
read and understand. Please rephrase.

Page 2, Paragraph 4: Please explain the initials cCOA, CDA and ROA.

Page 2, Example 1: Please give a pointer to Figure 1 in the first two lines of
the example.

Page 4, middle of the page, "the well-known ALC": The author should be very
careful when he uses expressions like this one. It is well known to whom?
Please consider a reader that does not know ALC.

Page 4, Section 2, "A constraint satisfaction ...": Please give references.

Page 4, Section 2, Paragraph 2, "An m-ary constraint ...": Please replace
"cdots" with "ldots" everywhere in the text.

Page 5: What the superscript b and t in expressions ``${\cal I}^b_U$''
and ``${\cal I}^t_U$'' stand for in your case. Please explain in advance. The notation used is
very "crowded". A simplification should be considered. This also applies for
Section 2.1-2.2 as well.

Page 5, Section 2.1, Paragraph 1: Please replace "verifying" with satisfying.

Page 6, Section 2.2, "P is k-consistent ...": Give examples.

Page 6, Section 2.2, "1-consistency, 2-consistency ...": What n in
n-consistency stands for. Please explain.

Page 6, Section 2.2, "A refinement": Please delete ", and".

Page 6, Section 2.3: This is a journal paper. It should contain a more thorough
presentation of related work. See also previous comments

Page 7, Items 1 and 2: You have explained only the symbols that appear in
Section 2. What are the other symbols (e.g., $\otimes$ and $\oplus$).

Section 3.1 and 3.2: I think that you should give representative names for the
calculi of [8] and [9] to distinguish between binary and ternary relations.
Then, you should use that name and the citation (e.g., the ternary relations
[9]) to refer to them in the text (and not just the citation). Using just the
citation number is confusing.

Section 3.1, last sentence: I thing that most people working on spatial
modelling would disagree with this statement. There many arguments that you can
say in favor of the projection-based model (for instance, it can be defined
using simply order constraint on the projections of the point on the x and y
axis) but I do not believe that it is more cognitive plausible that the cone
based model, on the contrary.

Section 4: Give more details. Give formal definitions. Give examples. Also
explain why you have adopted this model and not for instance the region based
models of [Goyal00, Billen04]

Section 4, Items 1 and 2: The model is defined very badly. Terms and relations
are not uniquely defined. For instance, relation ls is to the left of the
reference object but with respect to which reference frame. Also please mention
relation names instead of numbers in Fig. 4.

Section 4, last paragraph: Which are the relations of CDA and ROA. Please also
give examples and illustrations. "we omit brackets" this is the first time you
mention brackets.

Section 5.2, middle of page: What $\otimes$ stands for. Define and exemplify.

Section 5.2, Fig. 5: Use bigger font size. Does r1 runs in rows and r2 runs in
columns or vise versa? Please explain.

Page 11, Section 5.1: Please help the reader and present an illustration.

Section 5.2, Paragraph 1: The definition you have presented in this paragraph
is not the standard (existential/set-theoretic) composition but, it is
something weaker called consistency-based composition.
Set-theoretic/existential composition also requires that for two given regions
a and c such that $a r\_1\circ r\_2 b$ holds we can always find a region b such
that a r\_1 b and b r\_2 c. (For definitions please consider the following
references: [Bennett97, Ligozat01, Duentsch2000rarcc, Duentsch99a,
Skiadopoulos04]). For some models of spatial information composition and
consistency-based coincide. For other models they are different
[Duentsch2000rarcc, Duentsch99a, Skiadopoulos04]. Is consistency-based
composition that you discuss in this paper or (existential/set-theoretic)
composition? I believe that since you talk about algebras and calculi then the
answer should be (existential/set-theoretic) composition. Please clarify.

Section 5.2, Items 1 and 2. What $b_1$ and $b_2$ stand for? Please explain.

Page 12, Paragraph 1: Please present an illustration for every example.

Page 12, Paragraph 2: What is cCOA-CSP. The definition appears in Page 15. Move
it earlier.

Page 13, Figure 8: What $\bar{EQ}$ stands for?

Items 1 to 3 of Pages 12 and 13. I would like to see proof of these statements.

Page 15, Item 3: more? Explain.

Page 17, Theorem 1. This is not really a theorem. One can say that it is
actually a lemma. It only measures the computational complexity of the
procedure. The algorithm achieves 4-consistency and it runs in $O(n^4)$. No
surprise. I believe that the real theorem would have been to prove that
procedure PcS4c+ is complete. Can you prove that? This would be the major
contribution of the paper.

References. If you wish to abbreviate first names you should use a dot. Thus,
it is A. Isli and A. C. Cohn and not A Isli and A C Cohn respectively.

REFERENCES

[Bennett97]
   Bennett, B. and Isli, A. and Cohn, A. G.,
   ``When does a Composition  Table Provide a Complete and
                   Tractable Proof Procedure for a Relational Constraint Language?'',
   Proceedings of the IJCAI-97 workshop on Spatial and Temporal Reasoning,
   Nagoya, Japan,
   1997.

[Ligozat01]
   G. Ligozat,
   "When tables tell it all: qualitative spatial and
          temporal reasoning based on linear ordering",
   D.R. Montello (editor),
   volume 2205,
   LNCS,
   60--75,
   "Spatial Information Theory, proceedings of COSIT-01",
   2001,
   Springer.

[Duentsch2000rarcc],
  Ivo Duentsch and Hui Wang and Steve McCloskey,
  "A relation-algebraic approach to the
                  Region Connection Calculus",
  Theoretical Computer Science,
  2000.

[Duentsch99a]
  Ivo Duentsch and Hui Wang and Steve McCloskey,
  "Relation algebras in qualitative spatial reasoning",
  Fundamenta Informaticae,
  1999,
  volume 39,
  229--248.

[Skiadopoulos04]
    Skiadopoulos, S. and Koubarakis, M.,
    "Composing Cardinal Direction Relations",
    Artificial Intelligence,
    Elsevier,
    2004,
    152 (2),
    143-171.

[Skiadopoulos02]
    Skiadopoulos, S. and Koubarakis, M.,
    "Qualitative Spatial Reasoning with Cardinal Directions",
    Proceedings of the 7th International Conference on Principles
                  and Practice of Constraint Programming (CP'02),
    Van Hentenryck, P. (editor),
    Springer,
    LNCS 2470,
    341-355,
    Ithaca, New York,
    September,
    2002.

[Goyal00]
    Goyal, R. and Egenhofer, M.J.,
    ``Cardinal Directions Between Extended Spatial Objects",
    IEEE Transactions on Data and Knowledge Engineering
    (in press),
    2000.

[Billen04]
   Billen, B. and Clementini, E. ,
   "A Model for Ternary Projective Relations between Regions",
   LNCS 2992,
   310-328,
   Proceedings of the 9th Int'l Conference on Extending Database
                 Technology (EDBT'04),
   Bertino, E. and Christodoulakis, S. and Plexousakis, D. and
                 Christofides, V. and Koubarakis, M. and Boehm, K. and Ferrari, E. (editors),
   2004,
   Springer.

[Billen04b]
   Billen, B. and Clementini, E.,
   "Introducing a reasoning system based on ternary projective relations",
   Proceedings of SDH'04,
   2004.

\newpage\noindent
\begin{center}
THE COSIT'2001 NOTIFICATION LETTER\\
(as received on 2 May 2001)
\end{center}
Dear Dr. Isli,

Due to the large number of high-quality papers submitted to COSIT'01, 
the
Program Committee had a difficult task in selecting contributions for 
the
conference program.  In addition, issues of accessibility and topical
balance prevented us from accepting several papers of high quality.

I regret to inform you that your paper "Combining cardinal direction
relations with relative orientation relations in Qualitative Spatial
Reasoning" has not been selected for presentation at the Conference on
Spatial Information Theory (COSIT'01).  Below, please find reviewers'
comments you may find informative.  

We still enthusiastically invite you to consider attending COSIT'01, to 
be
held at the Inn at Morro Bay, near San Luis Obispo, September 19-23.

The COSIT'01 Web site is continually being updated.  You will soon find
information there for registering for the conference.  You will also 
find
there information on the Tutorial Program and the Doctoral Colloquium.  
The
site is:

	http://de.f125.mail.yahoo.com/ym/Compose?To=cosit01@geog.ucsb.edu\&YY=54185\\
		     \&order=down\&sort=date\&pos=2\&view=a\&head=b

The Program Committee and I thank you for your interest in COSIT'01 and
regret that the outcome was not more positive on this occasion.

Sincerely,

COSIT'01 Program Committee Chair

------------------PAPER  REVIEWS-------------------------------------

Paper Number:   [32]

Paper Title:    [Combining cardinal direction relations with relative
orientation relations in Qualitative Spatial Reasoning ]

1. Topic Appropriateness:  How appropriate is the paper to the themes 
of
COSIT?

[X] a. extremely appropriate

[] b. very appropriate

[] c. appropriate

[] d. only a little appropriate

[] e. not appropriate at all

[
The paper deals with combining two kinds of qualitative directional
information about the locations of points in 2D space:  one is absolute
information (related to a fixed frame of reference), and the other 
relative
information (where is a target object situated with respect to another,
relatively to a given point of view. Integrating various types of
qualitative spatial calculi is a valuable and "hot" topic (in 
particular
because each calculus, taken separately, is very weak). This work fits
nicely in this general  context by integrating two well-known calculi: 
the
cardinal direction calculus and (a coarse version of) Freksa's 
15-relation
calculus.
]

2. Scientific or technical quality of research:

[] a. Excellent

[X] b. Very good (upper 1/3)

[] c. Good (middle 1/3)

[] d. Fair (bottom 1/3)

[] e. Poor

Comments:
[The research is good research, the authors have an excellent knowledge 
of
the field and deal with it in a very competent way.]

3. How novel or innovative is the paper?

[] a. extremely innovative

[X] b. innovative

[] c. similar to other work but still somewhat innovative

[] d. not innovative

Comments:
[The paper is innovative: To my knowledge, the subject has not been 
treated
before, and the authors do quite a good job. However, for the time 
being,
they seem to have attained only modest results. The main contribution 
of
the paper consists in an algorithm which exploits some of the 
interaction
between relative and absolute orientations. But we do not know much 
about
what it achieves apart from 4-consistency and from the fact that it 
runs in
quartic time, which is fairly obvious.

This is not to imply that this work is of little worth. On the 
contrary, I
think that it is quite valuable and should be pursued. But it would be
quite good to have some more substantial results in the direction 
explored
by the authors. They should be encouraged to get them.]

4. Presentation (structure of the paper, language, graphical 
presentation,
etc.):

[] a. excellent

[X] b. good

[] c. acceptable with minor improvement [please detail]

[] d. needs major improvement [please detail]

Comments:
[
The presentation is basically quite good.
Some typos:

Page 7, l. -7 and -17:		"knowledge" instead of  "konwledge".

Page 14, references 14 and 15:	"2D" rather than "2d" (You may need "{D} 
in
the Latex source).
]

5. Disciplinary Breadth (content and presentation):

[X] a. will definitely appeal to more than one discipline

[] b. may appeal to more than one discipline

[] c. probably will appeal to only one discipline

[] d. will definitely appeal only to one discipline

Comments:
[
This is technical work, of course. But the elaboration of good 
qualitative
spatial calculi is
an important issue which has a definite import at least for AI, GIS 
theory,
linguistics, as well as an interest for computer science at large
(Constraint Satisfaction Problems). The authors make a reasonable 
effort
for justification.

6. Overall judgment: Do you believe that the paper should be included 
in
the program?

[] a. I STRONGLY recommend that the paper be included

[] b. I recommend that the paper be included

[X] c. I could go either way

[] d. I recommend that the paper NOT be included

[] e. I STRONGLY recommend that the paper NOT be included

7. How well do you know the topical area of the paper?

[] a. Extremely well, I consider myself an expert.

[X] b. Pretty well.

[] c. Moderately well, I am somewhat familiar with the area.

[] d. Not well, I'm really just guessing.

8. Other comments for the author(s)?

[Any dose of extra theoretical knowledge on the subject  would make the
paper definitively QUITE acceptable. ]

=================================END OF REVIEW 

Paper Number: [ 32 ]

Paper Title: [Combining cardinal direction relations with relative
orientation relations in Qualitative Spatial Reasoning ] 

1. Topic Appropriateness: How appropriate is the paper to the themes of 
COSIT? 

[] a. extremely appropriate 

[x] b. very appropriate 

[] c. appropriate 

[] d. only a little appropriate 

[] e. not appropriate at all 

Comments: The subject of the paper is spatial reasoning about direction 
and 
orientation relations. 

2. Scientific or technical quality of research: 

[] a. Excellent 

[] b. Very good (upper 1/3) 

[] c. Good (middle 1/3) 

[x] d. Fair (bottom 1/3) 

[] e. Poor 

Comments: The paper is overly complex and lacks any expansion of how 
the 
rather complex algebra has any relation to reality. Especially 
confusing is
the 
extension of binary relation calculus to ternary relation calculus.
Basically, a 
formal system in introduced and no rational description of how it may 
model 
reality is given. 

3. How novel or innovative is the paper? 

[] a. extremely innovative 

[] b. innovative 

[] c. similar to other work but still somewhat innovative 

[x] d. not innovative 

Comments: All I see is complexity. 

4. Presentation (structure of the paper, language, graphical 
presentation, 
etc.): 

[] a. excellent 

[] b. good 

[x] c. acceptable with minor improvement [please detail] 

[] d. needs major improvement [please detail] 

Comments: English is ill-structured and confusing at times. I suspect 
that it 
sounds better in the authors native language. For example, Section 1 
the
first 
sentence is: 
"Reasoning about orientation is one of the main aspects research 
 in Qualitative Spatial Reasoning (QSR) has focused on for about a 
decade 
now." This is arguably correct English, but the argument would take 
a while. Splitting prepositions is not considered good form. The 
structure is 
complex and no punctuation is used that might help to parse the 
sentence. A
more 
understandable version might be: 
 In the last decade, as one of its major topics, research in 
  Qualitative Spatial Reasoning (QSR) has focused on reasoning about 
orientation. Mixed with the complexity of the topic, the complexity 
of the language used makes the paper almost impossible to read. 

5. Disciplinary Breadth (content and presentation): 

[] a. will definitely appeal to more than one discipline 

[] b. may appeal 
to more than one discipline 

[] c. probably will appeal to only one 
discipline 

[x] d. will definitely appeal only to one discipline 

Comments: To be read by a general audience, this paper needs more 
information 
on its connection to reality. There is not sufficient support to show 
that 
solving a problem in the combined calculus would be logically 
equivalent to 
solving a problem in the real world. 

6. Overall judgment: Do you believe that the paper should be included 
in 
the program? 

[] a. I STRONGLY recommend that the paper be included 

[] b. I recommend 
that the paper be included 

[] c. I could go either way 

[] d. I 
recommend 
that the paper NOT be included 

[x] e. I STRONGLY recommend that the 
paper 
NOT be included 

7. How well do you know the topical area of the paper? 

[] a. Extremely well, I consider myself an expert. 

[] b. Pretty well. 

[x] c. Moderately well, I am somewhat familiar with the area. 

[] d. Not 
well, I'm really just guessing. 

8. Other comments for the author(s)? 
My reaction to this article is based mainly on its presentation. The 
paper is 
presented in such a complex fashion that it limits it own audience.
Published to 
a general audience, the paper will be read by only a few, and 
understood by
only 
a small portion of those. 

=================================END OF REVIEW

Paper Number:   [32   ]

Paper Title:    [Combining cardinal direction relations...   ]

1. Topic Appropriateness:  How appropriate is the paper to the themes 
of
COSIT?

[X] b. very appropriate

[DESCRIBE HOW IT FITS OR DOES NOT FIT - FREE TEXT]

2. Scientific or technical quality of research:

[X] a. Excellent

Comments:
[INSERT COMMENTS HERE, PARTICULARLY IF YOU SEE PROBLEMS - FREE TEXT]

3. How novel or innovative is the paper?

[X] b. innovative

Comments:
[INSERT COMMENTS HERE - FREE TEXT. LIST MISSING REFERENCES TO RELEVANT
LITERATURE HERE.]

4. Presentation (structure of the paper, language, graphical 
presentation,
etc.):

[X] a. excellent

Comments:
[INSERT COMMENTS HERE - FREE TEXT. PLEASE MENTION ANY TYPOS YOU FOUND]

5. Disciplinary Breadth (content and presentation):

[X] d. will definitely appeal only to one discipline

Comments:
[INSERT COMMENTS HERE - FREE TEXT. PLEASE MAKE RECOMMENDATIONS FOR
INCREASING ACCESSIBILITY TO OTHER DISCIPLINES, IF APPROPRIATE]

6. Overall judgment: Do you believe that the paper should be included 
in
the program?

[X] b. I recommend that the paper be included

7. How well do you know the topical area of the paper?

[X] c. Moderately well, I am somewhat familiar with the area.

8. Other comments for the author(s)?

[INSERT GENERAL COMMENTS HERE - FREE TEXT]

=================================END OF REVIEW
\end{document}